\documentclass[11pt]{article}

\usepackage[final]{acl}

\usepackage{times}
\usepackage{latexsym}
\usepackage[T1]{fontenc}
\usepackage[utf8]{inputenc}
\usepackage{microtype}
\usepackage{inconsolata}
\usepackage{graphicx}
\usepackage{amsmath}
\usepackage{amssymb}
\usepackage{booktabs}

\usepackage{tabularx}
\usepackage{multirow}
\usepackage{pifont}

\newcommand{\methodname}{SC-GRPO}

\usepackage{algorithm}
\usepackage{algpseudocode}
\usepackage[table]{xcolor}

\usepackage{listings}
\usepackage{tcolorbox}
\tcbuselibrary{listings,breakable}

\lstdefinestyle{promptstyle}{
    basicstyle=\ttfamily\small,
    breaklines=true,
    breakindent=0pt,
    showstringspaces=false,
    frame=none,
    numbers=none,
    keepspaces=true,
    columns=fullflexible,
    tabsize=2,
    escapeinside={(*}{*)},
}

\newtcolorbox{promptbox}[1][]{%
    colback=gray!5,
    colframe=gray!60,
    fonttitle=\bfseries\small,
    title={#1},
    breakable,
    boxrule=0.4pt,
    left=4pt, right=4pt, top=2pt, bottom=2pt,
}

\definecolor{phaseblue}{RGB}{30, 90, 160}
\definecolor{notecolor}{RGB}{130, 130, 130}
\definecolor{rowblue}{RGB}{220, 235, 252}

\NewDocumentCommand{\yingyu}
{ mO{} }{\textcolor{blue}{\textsuperscript{\textit{yingyu}}\textsf{\textbf{\small[#1]}}}}

\title{Learning from Own Solutions: Self-Conditioned Credit Assignment for Reinforcement Learning with Verifiable Rewards}

\author{%
Yingyu Shan\textsuperscript{1},
Yuhang Guo\textsuperscript{1, $\dagger$},
Zihao Cheng\textsuperscript{2},
Zeming Liu\textsuperscript{2},
Xiangrong Zhu\textsuperscript{3}, \\
\textbf{Xinyi Wang}\textsuperscript{3},
\textbf{Jiashu Yao}\textsuperscript{1},
\textbf{Wei Lin}\textsuperscript{3},
\textbf{Hongru Wang}\textsuperscript{3},
\textbf{Heyan Huang}\textsuperscript{1} \\
\textsuperscript{1}Beijing Institute of Technology \quad
\textsuperscript{2}Beihang University \quad
\textsuperscript{3}Independent Researcher \quad \\
\textsuperscript{$\dagger$}Corresponding author \quad
Email: \texttt{\{shanyingyu, guoyuhang\}@bit.edu.cn}
}

\begin{document}
\maketitle

\begin{abstract}

Reinforcement learning with verifiable rewards (RLVR) has driven
substantial progress in training LLMs for reasoning
tasks, but representative methods such as GRPO assign uniform credit
across all tokens, wasting gradient on routine tokens while
under-crediting pivotal reasoning steps.
Existing token-level credit assignment methods require resources beyond
the model's own rollouts. GRPO variants rely on process reward models
or ground-truth answers. Knowledge distillation assigns
credit through per-token divergence but requires
external teachers (On-Policy Distillation) or privileged information (On-Policy Self Distillation).
However, these dependencies limit
applicability in the pure RLVR setting.
We observe that conditioning the model on its own verified trajectories
induces a measurable per-token KL divergence between the original and
conditioned distributions, and prove that
distilling from a self-teacher constructed
by verified trajectories leads to infeasible
weighted-average solutions when multiple verified trajectories exist.
We propose \methodname{} (\textbf{S}elf-\textbf{C}onditioned GRPO), which uses KL divergence mentioned before as a multiplicative
weight on GRPO gradients. Across five benchmarks spanning math, code, and
agentic tasks, \methodname{} consistently outperforms 8.1\% over GRPO and 5.9\% over DAPO with 
stronger OOD performance. Moreover, \methodname{} achieves higher performance than OPD\footnote{Code is available at \url{https://github.com/BITHLP/SC-GRPO}}.

\end{abstract}

\section{Introduction}


\begin{figure}[ht]
    \centering
    \includegraphics[width=\linewidth]{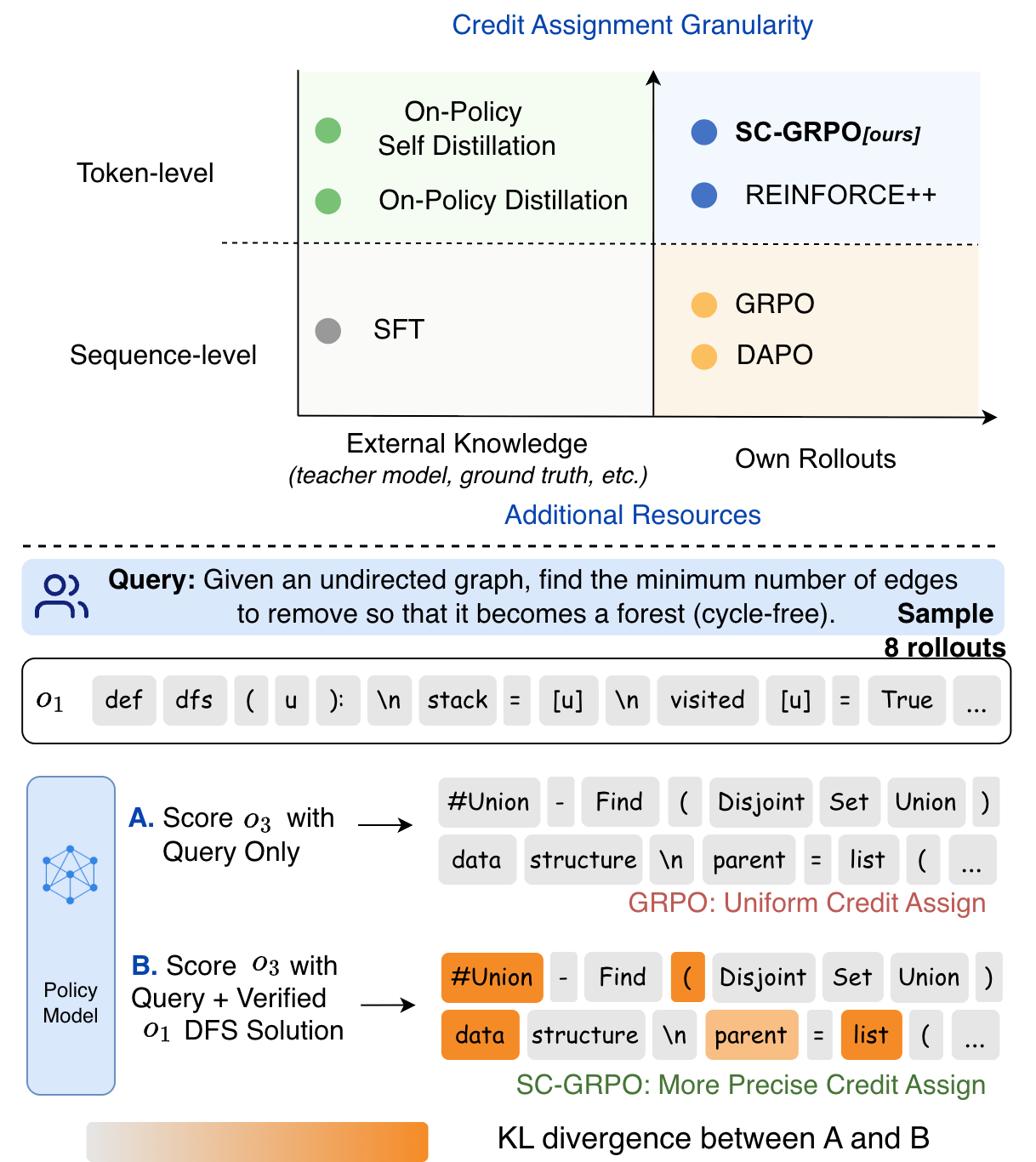}
    \caption{\textbf{Overview of \methodname{}.}
    \textit{Top:} positioning of \methodname{} among existing methods. 
    \textit{Bottom:} core mechanism illustrated on a LiveCodeBench example. 
    }
    \label{fig:teaser}
    \vspace{-2mm}
\end{figure}

Large language models (LLMs) have shown strong capabilities in various tasks, but competition-level problems remain
challenging. Reinforcement learning with verifiable rewards (RLVR)~\citep{shao2024deepseekmath,yu2026dapo, he2026searchr2, yao2025reff, cheng2026behavior2trippersonalizedtravelplanning} has driven
substantial progress in tackling such problems. Representative methods in RLVR including GRPO~\cite{shao2024deepseekmath, liuunderstanding, zheng2025group}
assign a single scalar reward per rollout, so every token shares the
same advantage. This uniform credit cannot identify which tokens
caused success or failure, diluting gradient across routine tokens
while under-crediting pivotal reasoning
steps~\citep{he2026rethinking,xie2025unlocking}.


To enable token-level credit assignment, existing approaches require
resources beyond the model's own rollouts (Figure~\ref{fig:teaser}, top). Recent GRPO variants have
explored different directions: some train process reward models on
step-level annotations~\citep{wang2024math, cui2025process}, while others rely on ground-truth
answers~\citep{wang2025information}. 
Alternatively, knowledge distillation directly assigns credit at the
token level through per-position divergence between student and teacher
distributions \cite{hinton2015distilling}.
On-Policy Distillation (OPD)~\citep{agarwal2024gkd,kodistillm} uses a stronger
external teacher to provide token-level supervision, and On-Policy
Self-Distillation (OPSD)~\citep{hubotter2025sdpo,zhao2025opsd}
similarly requires privileged information. These dependencies limit
applicability in the pure RLVR setting, where only a binary
verifier is available.

However, we can identify which tokens carry credit by comparing
predictions with and without access to the model's own verified solution.
Recent work has shown that conditioning models on their own verified
trajectories improves generation quality~\citep{yao2026incorporating}.
Building on this observation, we propose \methodname{} (\textbf{S}elf-\textbf{C}onditioned GRPO), which uses
such conditioning to induce a measurable per-token
distributional shift as a credit assignment signal.
As shown in Figure~\ref{fig:teaser} bottom, 
for problems where the model produces both correct and incorrect rollouts,
\methodname{} injects one verified correct solution as an in-context demonstration and re-scores all
remaining rollouts under this prompt, constructing a self-conditioned teacher. 
A complete version of this example is provided in Figure~\ref{fig:case-study}.

We quantify this shift via per-token Kullback--Leibler (KL) divergence
between the original and demonstration-conditioned next-token distributions.
Inspired by OPSD~\citep{zhao2025opsd}, a natural approach would be to add this KL as an auxiliary
distillation loss. However, when multiple distinct
verified trajectories exist, the loss-minimizing student converges to a
weighted average over these teachers, a distribution that does not
correspond to any feasible trajectory (we formalize this in \S\ref{sec:opsd-fails}).
Empirically, such additive formulations consistently fail to improve over GRPO
(\S\ref{sec:kl-weight-vs-loss}).

Instead, \methodname{} uses the KL signal purely as a multiplicative
weight on the GRPO gradient: the reward determines the update direction,
and the KL determines only its intensity at each token. The same
self-conditioned teacher handles both partial-solve groups, where a
verified correct trajectory serves as the reference and the KL weight
sharpens credit assignment, and solve-none groups, where a random rollout
serves as the reference and the KL signal provides a slight exploration bonus that encourages diversity. 
On five benchmarks spanning math, code, and multi-turn
agentic tasks, \methodname{} improves Average@8 by 8.1\% over GRPO and
5.9\% over DAPO, while achieving more consistent and higher performance than OPD
with external teachers. 

Our contributions are as follows:
\begin{itemize}
\item In RLVR, verified trajectories are the only supervision beyond the
  binary verifier. We prove that distilling from a teacher constructed
  using these trajectories leads to infeasible solutions, 
  and demonstrate empirically that such formulations
  fail to improve over GRPO.

\item We propose \methodname{}, which constructs a teacher by conditioning
  the model on its own verified trajectories and uses the resulting
  token-level KL divergence as a multiplicative weight on GRPO gradients.

\item Across five benchmarks spanning math, code, and agentic tasks,
  \methodname{} consistently outperforms RL baselines 
  and achieves higher and more stable
  performance than OPD.

\end{itemize}

\section{Related Work}

\subsection{On-Policy Distillation}

On-policy distillation trains on student generated rollouts with a divergence to the teacher distribution \citep{agarwal2024gkd,guminiplm,kodistillm,shingtaid}, eliminating the train-test mismatch of offline distillation \citep{hinton2015distilling}.
The divergence objective has been refined through skew-KL \citep{kodistillm}, $\alpha$-$\beta$-divergence \citep{wang2025abkd}, and adaptive interpolation \citep{shingtaid}, highlighting the sensitivity of letting divergence set the gradient direction. Moreover, all variants require a stronger external teacher.

Self-distillation removes the external teacher dependency by constructing a teacher from the model itself, typically conditioned on privileged information such as textual feedback \citep{hubotter2025sdpo}, ground truth \citep{zhao2025opsd,ding2026hdpo, yang2026selfdistilledrlvr}, or expert demonstrations \citep{shenfeld2026selfdistillation}.
\methodname{} uses KL as a per-token weight, eliminating sensitivity to divergence form choice, and conditions the teacher solely on the model's own verified rollouts.

\subsection{Token-Level Credit Assignment in RL}

Process reward models (PRMs) \citep{lightman2023letsverifystepstep, wang2024math, ArmoRM} provide step-level supervision but require training a separate reward model with large-scale annotation, and remain vulnerable to reward hacking \citep{juneja2025adversarial,gao2024designing}.
Annotation-free alternatives estimate token values through Monte Carlo sampling \citep{guanrstar} or segment-level sampling \citep{guo2026segment}, requiring additional rollout compute.
GRPO-based methods also need external PRM models \cite{cui2025process} or ground truth \cite{wang2025information} to gain token-level credit during training.
In contrast, \methodname{} only depends on the model's own rollouts to obtain token-level reward without external resources. 


\section{Preliminaries}
\label{sec:preliminaries}

\paragraph{Notation}
Let $\pi_\theta$ denote the policy parameterized by $\theta$.
Given a query $x$ sampled from the training dataset $\mathcal{D}$, the policy
generates a response $o=(o_1,\ldots,o_{|o|})\sim\pi_\theta(\cdot\mid x)$,
where $o_t$ is the token at position $t$ and $o_{<t}$ denotes the prefix before
position $t$. Each response is evaluated by a verifier $r$,
yielding a scalar reward $r(x,o) \in [0, 1]$.

\subsection{GRPO}
For each query $x$, GRPO samples a group of $G$ responses
$\{o_i\}_{i=1}^{G}$ from the old policy $\pi_{\theta_{\mathrm{old}}}$. Each response is evaluated
by the verifier, yielding $r_i=r(x,o_i)$. Instead of learning a value
function, GRPO normalizes rewards across the $G$ responses in the same group
to obtain a sequence-level advantage. Equivalently, every token position $t$ in
response $o_i$ shares the same advantage:
\begin{equation}
\hat{A}_{i,t} = \hat{A}_i
= \frac{r(x,o_i) - \mathrm{mean}(\{r(x,o_k)\}_{k=1}^{G})}{
\mathrm{std}(\{r(x,o_k)\}_{k=1}^{G})}.
\label{eq:grpo-adv}
\end{equation}
The policy is then updated with following objective over response
tokens:


\begin{align*}
& \mathcal{J}_{\mathrm{GRPO}}(\theta)
= \mathbb{E}_{x\sim\mathcal{D},\, \{o_i\}_{i=1}^{G}\sim\pi_{\theta_{\mathrm{old}}}(\cdot\mid x)}
\Bigg[
\frac{1}{G}\sum_{i=1}^{G} 
\frac{1}{|o_i|} \\
& \sum_{t=1}^{|o_i|} 
\min\!\bigl(
  \rho_{i,t}\hat{A}_{i,t},\,
  \mathrm{clip}(\rho_{i,t}, 1-\epsilon, 1+\epsilon)\hat{A}_{i,t}
\bigr)
\Bigg], \\
& \mathrm{with} \quad \rho_{i,t}=
\frac{\pi_\theta(o_{i,t} \mid x, o_{i,<t})}{
\pi_{\theta_{\mathrm{old}}}(o_{i,t} \mid x, o_{i,<t})}.
\end{align*}


\subsection{On-Policy Distillation}

OPD \cite{agarwal2024gkd} provides dense token-level signal along the student's own trajectories. 

Let $\pi_\theta$ and $\pi_T$ denote the student and frozen teacher policies, 
respectively. Given a prompt $x \sim \mathcal{D}$, the student samples a 
response $o \sim \pi_\theta(\cdot \mid x)$, which induces prefix states 
$s_t = (x, o_{<t})$. At each prefix, the student and teacher next-token 
distributions are $\pi_\theta(\cdot \mid s_t)$ and $\pi_T(\cdot \mid s_t)$, 
and the token-level OPD objective is
\begin{equation}
\begin{aligned}
\mathcal{L}_{\mathrm{OPD}}(\theta)
&= \mathbb{E}_{\,x \sim \mathcal{D},\; o \sim \pi_\theta(\cdot \mid x)}
\Big[
\frac{1}{|o|}\sum_{t=1}^{|o|}
\\
& \quad D\!\bigl(\pi_T(\cdot \mid s_t)\,\|\,\pi_\theta(\cdot \mid s_t)\bigr)
\Big],
\end{aligned}
\label{eq:opd}
\end{equation}
where $D$ is a divergence such as the Kullback--Leibler (KL) divergence or 
the Jensen--Shannon divergence (JSD).

OPSD \cite{ding2026hdpo, hubotter2025sdpo, zhao2025opsd} removes the requirement of external teacher by conditioning the same model on privileged information $c$ (e.g.,
ground-truth or expert demonstration). The student distribution remains $\pi_\theta(\cdot \mid s_t)$, while the self-teacher is instantiated as:
\begin{equation}
\widetilde{\pi}_\theta(\cdot \mid s_t, c)
\;:=\;
\operatorname{sg}\!\left[
\pi_\theta(\cdot \mid x, c, o_{<t})
\right],
\label{eq:self-teacher}
\end{equation}
where $\operatorname{sg}[\cdot]$ denotes the stop-gradient operator, 
indicating that the teacher distribution is treated as a fixed target for 
the current update. The training objective keeps the same form as 
Eq.~\eqref{eq:opd}, with $\pi_T(\cdot \mid s_t)$ replaced by 
$\widetilde{\pi}_\theta(\cdot \mid s_t, c)$.

\section{Why Self-Distillation Fails in RLVR}
\subsection{Available Supervision in RLVR}

For a query $x$, GRPO samples a group of responses 
$\{o_i\}_{i=1}^{G} \sim \pi_{\theta_{\mathrm{old}}}(\cdot \mid x)$, and a 
verifier assigns a sequence-level reward $r_i = r(x, o_i) \in \{0, 1\}$ to 
each response. Let
\begin{equation}
\mathcal{C}(x) = \{\, o_i : r(x, o_i) = 1,\ i \in \{1, \dots, G\} \,\}
\notag
\end{equation}
denote the set of \emph{verifier-approved} trajectories.

In conventional self-distillation mentioned before, the teacher is conditioned on some form of
privileged information. In the RLVR setting, however, no such
privileged information is available. Beyond the query $x$ and the scalar reward
$r$, the only additional information we can access is a
verifier-approved trajectory drawn from the model's own rollouts, i.e., a sample
from $\mathcal{C}(x)$.
\textbf{A verified trajectory is not privileged information.} It certifies
only end-task correctness, not reasoning quality. The trajectory may contain
exploratory steps, redundant derivations, or recoverable mistakes, and when
multiple trajectories in $\mathcal{C}(x)$ succeed, they may follow
contradictory reasoning paths.

\subsection{Why OPSD Fails in RLVR}
\label{sec:opsd-fails}

Despite this limitation, verified trajectories are the only additional
structure available in RLVR. A natural but naive adaptation is therefore
to treat them as a substitute for privileged context.
Concretely, for a 
student rollout $o_i$ with prefix states 
$s_{i,t} = (x, o_{i,<t})$ and a verifier-approved 
trajectory $\tau_j \in \mathcal{C}(x)$, we
instantiate the generic OPSD formulation in
Section~\ref{sec:preliminaries} by taking $\tau_j$
as the privileged context:
\begin{equation}
\widetilde{\pi}_\theta(\cdot \mid s_{i,t},\, \tau_j)
\;:=\;
\operatorname{sg}\!\left[
\pi_\theta(\cdot \mid x,\, \tau_j,\, o_{i,<t})
\right].
\label{eq:self-teacher-rlvr}
\end{equation}
The training objective keeps the same form as
Eq.~\eqref{eq:opd}, replacing the external teacher
$\pi_T$ with the self-conditioned teacher
$\widetilde{\pi}_\theta(\cdot \mid s_{i,t},\, \tau_j)$.

This adaptation, however, does not work. A pointwise 
analysis (detailed in Appendix~\ref{app:opsd-derivation}) 
shows that under both forward and reverse KL, the 
loss-minimizing student distribution at any prefix 
$s_{i,t}$ is an \textbf{aggregate} of the 
self-conditioned teachers $\{p_j\}$ across 
$\mathcal{C}(x)$:
\begin{equation}
\!\!q^\star \propto
\begin{cases}
\sum_{j} \mu_x(j)\, p_j
& \text{(Forward KL),} \\[2pt]
\prod_j p_j^{\,\mu_x(j)}
& \text{(Reverse KL),}
\end{cases}
\notag
\end{equation}
where $p_j := \widetilde{\pi}_\theta(\cdot \mid s_{i,t}, \tau_j)$
and $\mu_x$ is the selection rule over
$\mathcal{C}(x)$.
This average distribution is problematic:
(i)~it need not correspond to any single feasible
trajectory; (ii)~it assigns equal weight to all prefixes,
even those where verified trajectories disagree about
the next token; and (iii)~it inherits whatever each $\tau_j$ happens to
take to reach the correct answer. These issues are fundamental to any distillation-based
approach that treats the teacher as a target distribution.

\paragraph{Empirical validation}
The analysis above characterizes the pointwise optimization tendency
of direct OPSD objectives. In Section~\ref{sec:kl-weight-vs-loss},
we examine its practical consequence using five additive distillation
variants constructed from the same self-conditioned teacher. All five
variants underperform GRPO and \methodname, empirically supporting the failure mode
identified by our analysis.
\begin{figure*}[t]
    \centering
    \includegraphics[width=\linewidth]{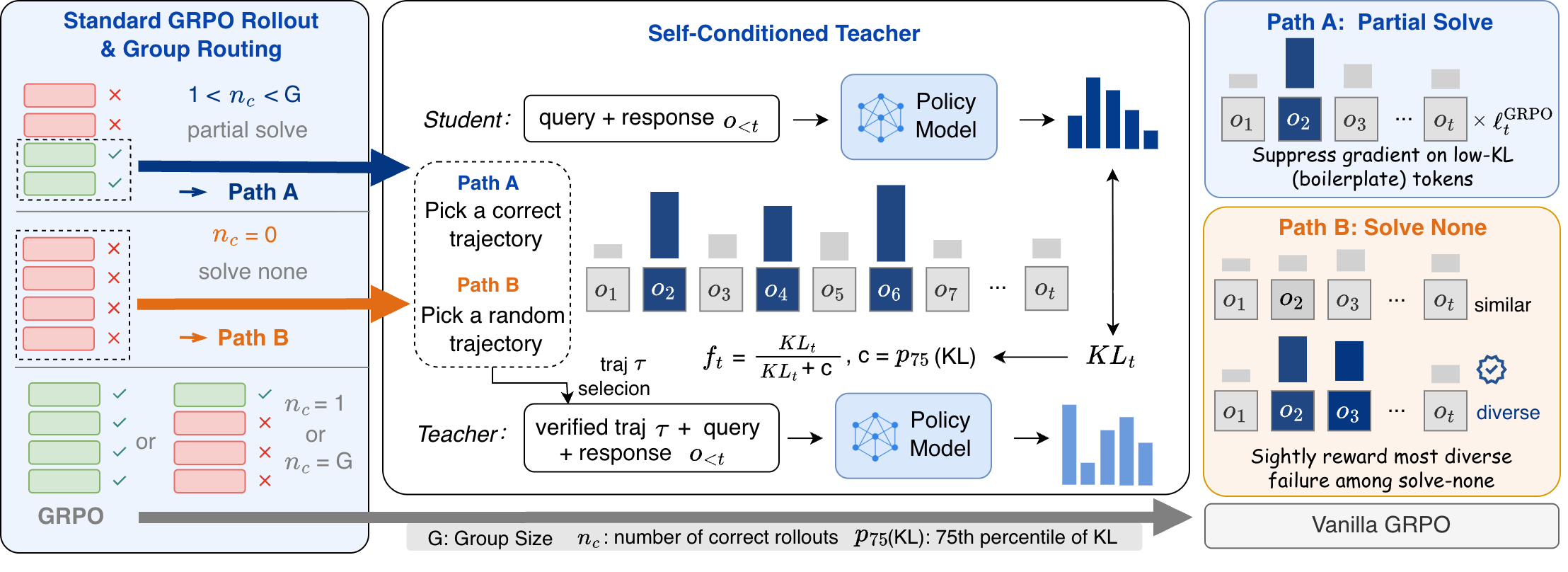}
    \caption{\textbf{Overview of \methodname{}.}
    We construct a self-conditioned teacher by conditioning the model on a reference
    trajectory, compute token-level KL divergence, and use it to weight the GRPO gradient.
    \label{fig:method}}
    \vspace{-2mm}
\end{figure*}

\section{\methodname{}}
\label{sec:method}
\definecolor{grpocom}{RGB}{120,130,155}
\definecolor{kwcom}{RGB}{30,130,120}
\definecolor{divcom}{RGB}{140,90,30}

\begin{algorithm}[t]
\caption{\methodname{}}
\label{alg:sc-grpo}
\small
\begin{algorithmic}[1]
\Require policy $\pi_\theta$; query $x$; group size $G$; diversity coefficient $\alpha$
\vspace{3pt}

\Statex \textcolor{grpocom}{\textbf{// Phase 1: Standard GRPO rollout}}
\State Sample $G$ responses $\{o_1,\ldots,o_G\}\sim\pi_\theta(\cdot\mid x)$
\State Verify each $o_i$; let $n_c$ = number of correct responses
\State Compute group-normalized advantage $\hat{A}_i$ for each $o_i$
\vspace{3pt}

\Statex \textcolor{kwcom}{\textbf{// Phase 2: KL signal (if $2 \le n_c < G$)}}
\Statex \textit{Compute all $D_t$ and $f_t$ in Phases 2--3 without gradient tracking.}
\For{each response $o_i$}
  \State Pick a correct response $\tau \ne o_i$ as demonstration
  \State Run teacher forward: $\widetilde{\pi}_\theta(\cdot\mid x, \tau, o_{i,<t})$
  \State Run student forward: $\pi_\theta(\cdot\mid x, o_{i,<t})$
  \State $D_t \leftarrow \mathrm{KL}( \widetilde{\pi}_\theta \| \pi_\theta)$ at each token $t$
\EndFor
\State $c \leftarrow \max\big(p_{75}(\{D_t\}_{\text{all tokens}}),\; 10^{-4}\big)$
\State $f_t \leftarrow D_t / (D_t + c)$ \hfill{\color{kwcom}// per-token weight $\in [0,1)$}
\vspace{3pt}

\Statex \textcolor{divcom}{\textbf{// Phase 3: Diversity signal (if $n_c = 0$)}}
\State Pick a random response $o_r$ as reference
\For{each $o_i \ne o_r$}
  \State Teacher forward with $o_r$ in context (same as Phase 2)
  \State Compute $D_t, f_t$ as in Phase 2
  \State $s_i \leftarrow \frac{1}{|o_i|}\sum_t f_t$ \hfill{\color{divcom}// diversity score}
\EndFor
\State $\hat{A}_i \leftarrow \alpha \cdot (s_i - \bar{s}) / \mathrm{std}(s)$ for each $o_i$
       \hfill{\color{divcom}// exploration signal for zero-reward groups}
\vspace{3pt}

\Statex \textcolor{grpocom}{\textbf{// Phase 4: Weighted update}}
\State Update $\theta$ to maximize
       $\displaystyle\frac{1}{G}\sum_i\frac{1}{|o_i|}\sum_t f_t \cdot \ell^{\mathrm{GRPO}}_{i,t}(\hat{A}_i)$
\Statex \hfill{\color{grpocom}// $f_t{=}1$ for groups skipping Phase 2\&3}
\end{algorithmic}
\end{algorithm}

\paragraph{Overview}
Section~\ref{sec:opsd-fails} shows that directly using verified trajectories as
OPSD distillation targets fails in RLVR.
Therefore, we propose \methodname{} (\textbf{S}elf-\textbf{C}onditioned GRPO), which takes a different approach.
We construct a \emph{self-conditioned teacher} by conditioning the model on a
verified trajectory, then measure the per-token distributional shift between
the teacher and the original student.

This shift, quantified as token-level KL divergence, identifies which tokens
depend on access to a verified solution. Tokens with small KL make the same prediction
regardless of whether a verified solution is available.
The token's choice is unrelated to the
rollout's outcome, so applying sequence-level credit here misattributes
success or failure.
Tokens with large KL depend on verified solution, so the reward
signal is informative. \methodname{} uses this KL purely as a multiplicative
weight on the GRPO gradient: the reward determines the update direction, and
the KL determines its intensity at each token.

Figure~\ref{fig:method} and Algorithm~\ref{alg:sc-grpo} illustrate the method.
The core mechanism is described in \S\ref{sec:sc-grpo}: we construct a
self-conditioned teacher by conditioning the model on a reference trajectory,
compute token-level KL, and use it to weight the GRPO gradient. The reference
is selected based on the number of correct rollouts $n_c$ in the group: verified trajectories for
partial-solve groups ($2 \le n_c < G$), random rollouts for solve-none groups
($n_c{=}0$), and standard GRPO for $n_c{=}1$ or $n_c{=}G$.
Section~\ref{sec:overhead} analyzes the computational overhead.

\subsection{Self-Conditioned GRPO}
\label{sec:sc-grpo}


\paragraph{Teacher construction}
For a rollout $o_i$ with prefix states $s_{i,t} = (x, o_{i,<t})$ and a
verified trajectory $\tau \in \mathcal{C}(x)$ , we instantiate the self-conditioned teacher
$\widetilde{\pi}_\theta(\cdot \mid s_{i,t}, \tau)$ via
Eq.~\eqref{eq:self-teacher-rlvr}, where $\tau$ appears only in the teacher's
system prompt. Appendix~\ref{app:prompts} provide the complete prompt templates. Both teacher and student are scored on the same prefix $o_{i,<t}$
of $o_i$. At each response token, we compute the forward KL from teacher to student:
\begin{equation}
D_{i,t}
=
\operatorname{sg}\!\left[
\mathrm{KL}\!\left(
\widetilde{\pi}_\theta(\cdot \mid s_{i,t}, \tau)
\;\middle\|\;
\pi_\theta(\cdot \mid s_{i,t})
\right)
\right],
\notag
\end{equation}

Both the teacher and student distributions used to compute
$D_{i,t}$ are evaluated without gradient tracking. 
The KL signal affects only the
magnitude of each token update and does not introduce an additional
distillation gradient.

\paragraph{KL weighting}
We map each $D_{i,t}$ to a bounded weight $f(D_{i,t}) \in [0,1)$ to modulate
the GRPO gradient: tokens with small KL are downweighted,
while tokens with large KL are preserved.
We map each $D_{i,t}$ to a bounded weight via
\begin{equation}
\begin{aligned}
f(D_{i,t})&=\frac{D_{i,t}}{D_{i,t}+c},\\
c&=\max\!\left(
P_{75}(\mathcal{D}_{\mathrm{act}}),\, c_{\min}
\right),
\end{aligned}
\notag
\end{equation}
where $\mathcal{D}_{\mathrm{act}}$ is the set of active token KL values in the
current micro-batch, $P_{75}$ denotes its 75th percentile, and $c_{\min}>0$
is a small floor. We use $P_{75}$ instead of median because token-level KL is sparse
(most tokens have near-zero KL). $P_{75}$ ensures low-KL tokens are suppressed while
high-KL tokens are preserved. 
We visualize and analyze the KL distribution and weighting function in Appendix~\ref{app:kl_weighting} and Appendix~\ref{app:kl_distribution}.
A detailed case study is provided in Appendix~\ref{app:case-study}.

\paragraph{Group routing}
The reference trajectory $\tau$ and advantage $\hat{A}_i$ are selected based on
the group's solve pattern. Partial-solve groups have both correct and incorrect
rollouts, allowing us to use verified trajectories as references. Solve-none
groups have zero GRPO gradient, so we extract exploration signal via diversity.

Partial-solve groups ($2 \le n_c < G$):
For each rollout $o_i$, we sample a verified trajectory $\tau \in \mathcal{C}(x)$
uniformly (with $\tau \ne o_i$ when $o_i \in \mathcal{C}(x)$). We use the
GRPO advantage $\hat{A}_i$ in Eq.~\eqref{eq:grpo-adv}.

Solve-none groups ($n_c = 0$):
We sample one rollout $o_r$ uniformly as reference. For each remaining rollout
$o_i$ ($i \ne r$), we compute KL against the reference-conditioned teacher.
The average KL weight over the reference-length prefix becomes a diversity score:
\begin{equation}
s_i
=
\frac{1}{L_i}
\sum_{t=1}^{L_i} f(D_{i,t}),
\quad
L_i=\min(|o_i|,|o_r|),
\notag
\end{equation}
where $L_i$ truncates at the reference length. Since the
reference itself is a failed trajectory, we apply stricter supervision to prevent
reward hacking: tokens beyond the reference would have artificially high KL and
inflate diversity scores for longer rollouts. Scores are normalized within the
group to form a pseudo-advantage:
\begin{equation}
\hat{A}^{\mathrm{div}}_i
=
\frac{s_i-\mathrm{mean}(\{s_k\}_{k=1}^{G})}
{\mathrm{std}(\{s_k\}_{k=1}^{G})+\epsilon},
\label{eq:diversity-adv}
\end{equation}
which replaces $\hat{A}_i$ with $\alpha \hat{A}^{\mathrm{div}}_i$ in the \methodname{} objective. 
Since all rollouts failed which bring zero GRPO advantage, this diversity signal encourages exploring new modes
rather than repeating common ones.

Fallback ($n_c = 1$ or $n_c = G$):
Standard GRPO (no KL weighting).

\methodname{} optimizes the weighted GRPO objective:
\begin{equation}
\begin{split}
&J_{\mathrm{SC\text{-}GRPO}}(\theta)
= \mathbb{E}\Big[\tfrac{1}{G}\!\sum_{i=1}^{G}\tfrac{1}{|o_i|}\!\sum_{t=1}^{|o_i|}\operatorname{sg}\!\left[f(D_{i,t})\right]\, \\
&\min\!\big(
\rho_{i,t}\hat{A}_i,\,
\mathrm{clip}(\rho_{i,t},1{-}\epsilon,1{+}\epsilon)\hat{A}_i
\big)\Big],
\end{split}
\notag
\vspace{-2mm}
\end{equation}
where $\rho_{i,t}=\pi_\theta(o_{i,t}\mid s_{i,t})/\pi_{\theta_{\mathrm{old}}}(o_{i,t}\mid s_{i,t})$
is the importance ratio, $\hat{A}_i$ is the advantage (GRPO advantage from
Eq.~\eqref{eq:grpo-adv} for partial-solve groups, $\alpha \hat{A}^{\mathrm{div}}_i$ from Eq.~\eqref{eq:diversity-adv}
for solve-none groups). The KL weight $f(D_{i,t})$ is applied symmetrically: both correct and incorrect
rollouts are downweighted at low-KL tokens, regardless of their advantage sign.

\begin{figure}[ht]
    \centering
    \includegraphics[width=\linewidth]{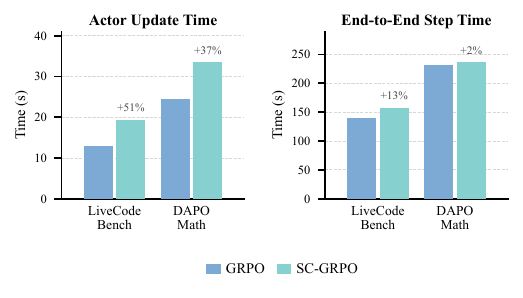}
    \caption{\textbf{Timing overhead of \methodname{} vs.\ GRPO.}
    Mean per-step wall-clock time (seconds) across training steps.}
    \label{fig:timing}
    \vspace{-2mm}
\end{figure}

\begin{table*}[t]
\centering
\setlength{\tabcolsep}{4pt}
\renewcommand{\arraystretch}{1.3}
\resizebox{\textwidth}{!}{%
\begin{tabular}{l *{12}{c}}
\toprule[0.08em]
\multirow{2}{*}{\textbf{Method}}
& \multicolumn{2}{c}{\textbf{AIME24}}
& \multicolumn{2}{c}{\textbf{AIME25}}
& \multicolumn{2}{c}{\textbf{LCB v6}}
& \multicolumn{2}{c}{\textbf{AppWorld}}
& \multicolumn{2}{c}{\textbf{WebShop}}
& \multicolumn{2}{c}{\textbf{Avg.}} \\
\cmidrule(lr){2-3}
\cmidrule(lr){4-5}
\cmidrule(lr){6-7}
\cmidrule(lr){8-9}
\cmidrule(lr){10-11}
\cmidrule(lr){12-13}
& \textbf{Avg@8} & \textbf{Pass@8}
& \textbf{Avg@8} & \textbf{Pass@8}
& \textbf{Avg@8} & \textbf{Pass@8}
& \textbf{Avg@8} & \textbf{Pass@8}
& \textbf{Avg@8} & \textbf{Pass@8}
& \textbf{Avg@8} & \textbf{Pass@8} \\
\midrule[0.05em]

Qwen3-8B
& 25.41 & 46.67 & 18.75 & 30.00 & 26.33 & 35.11 & 9.65 & 26.31 & 8.13 & 14.00 & 17.65 & 30.42 \\
\rowcolor{rowblue}
\multicolumn{13}{c}{\textit{\textbf{Reinforcement Learning}}} \\
GRPO & 41.67 & \underline{70.00} & 34.16 & 50.00 & 39.59 & 43.51 & 12.50 & 33.33 & 68.93 & 69.40 & 39.37 & 53.25 \\
DAPO
& \underline{43.75} & \underline{70.00} & \underline{37.91} & \underline{53.33} & \underline{40.74} & \underline{44.27} & \underline{13.99} & 33.33 & \underline{71.50} & \underline{75.00} & \underline{41.58} & \underline{55.19} \\
REINFORCE++$^\ddagger$ & 18.33 & 26.67 & 16.25 & 30.00 & 39.69 & \underline{44.27} & -- & -- & -- & -- & -- & -- \\
\rowcolor{rowblue}
\multicolumn{13}{c}{\textit{\textbf{On-Policy Self Distillation with External Demonstrations}}} \\
MiniMax-M2.7$^\dagger$
& 29.17 & 50.00 & 19.58 & 36.67 & 26.05 & 35.88 & 9.87 & \underline{35.08} & 11.40 & 20.00 & 19.21 & 35.53 \\
DeepSeekv4-Pro$^\dagger$
& 33.75 & 56.67 & 22.08 & 33.33 & 26.05 & 35.88 & 9.43 & 29.82 & 9.88 & 16.80 & 20.24 & 34.50 \\
Oracle
& 33.75 & 56.67 & 22.08 & 33.33 & 25.38 & 36.64 & 9.43 & 31.57 & 10.38 & 21.60 & 20.20 & 35.96 \\
\rowcolor{rowblue}
\multicolumn{13}{c}{\textit{\textbf{Ours}}} \\
\methodname{}
& \textbf{51.67} & \textbf{80.00} & \textbf{42.08} & \textbf{63.33} & \textbf{43.22} & \textbf{48.85} & \textbf{22.36} & \textbf{49.12} & \textbf{77.88} & \textbf{79.25} & \textbf{47.44} & \textbf{64.11} \\
\bottomrule[0.08em]
\end{tabular}
}
\vspace{-2mm}
\caption{\textbf{Main results.} We report Avg@8 and Pass@8 across five RLVR benchmarks.
$^\dagger$ External LLM that generates demonstration trajectories for OPSD.
Oracle collects as many correct demonstrations as possible by extending the thinking budget.
\textbf{Bold} indicates the best result, \underline{underline} indicates the second-best. 
$^\ddagger$ REINFORCE++ experienced training collapse on AppWorld and WebShop. 
Detailed training
dynamics are provided in Appendix~\ref{app:reinforcepp_dynamics}.
\label{tab:main-results}}
\vspace{-2mm}
\end{table*}

\subsection{Computational Overhead}
\label{sec:overhead}

\methodname{} adds a single new operation compared to GRPO: a teacher forward pass for per-token KL weighting. As shown in Figure~\ref{fig:timing}, this makes the actor
update 51\% slower on LiveCodeBench and 37\% slower on DAPO-Math, but end-to-end step time grows by only 13\% and 2\% respectively, since rollout generation dominates total
latency. Combined with the gains in Section~\ref{sec:main-results}, this confirms that \methodname{}'s improvements come from better use of existing rollouts rather than
additional compute.

\section{Experiments}
\label{sec:experiments}

\subsection{Setup}

\paragraph{Tasks}
Following \citet{cheng2026mem2evolve}, we train and evaluate on RLVR tasks spanning mathematical
reasoning, code generation, and multi-turn agentic interaction.
(\textit{i})~\textbf{Math}: we train on DAPO-Math-17k~\citep{yu2026dapo} and evaluate on 
AIME~2024 \& 2025.
(\textit{ii})~\textbf{Code}: we use
LiveCodeBench~v6~\citep{jain2025livecodebench} (LCB), holding out
half of each problem's unit tests for training and using the rest
for evaluation.
(\textit{iii})~\textbf{AppWorld}~\citep{trivedi2024appworld}, 
(\textit{iv})~\textbf{WebShop}~\citep{yao2022webshop}. For both AppWorld and WebShop, we train
and evaluate on the official splits.
Except for the separate OPD comparison in Figure~\ref{fig:opd}, all
runs use the same \textbf{Qwen3-8B base model with thinking mode disabled}.
For Figure~\ref{fig:opd}, we follow the original training and inference
configuration of \citet{li2026rethinkingOPD}.
Full per-task settings are detailed in Appendix~\ref{app:training_details}.

\paragraph{Metrics}
For each query we sample eight trajectories and report two
metrics: \textbf{Avg@8}, the mean verifier reward
across the eight samples; and \textbf{Pass@8}, the fraction of queries solved by
at least one of the eight samples.

\subsection{Baselines}

We compare \methodname{} against two families of baselines, all
sharing the same base model, training data, rollouts-per-prompt, and
step budget as our method.

\paragraph{Reinforcement learning}
\textbf{GRPO}~\citep{shao2024deepseekmath} is the canonical RLVR
algorithm. \textbf{DAPO}~\citep{yu2026dapo} is a stronger GRPO
variant incorporating decoupled clipping, dynamic sampling, and
token-level loss normalization. 
\textbf{REINFORCE++}~\citep{hu2025reinforcepp}
simplifies PPO by removing the critic and shaping rewards with
per-token KL penalties.

\paragraph{OPSD with external demonstrations}
Following \citet{zhao2025opsd}, we report OPSD baselines that
supply the teacher with demonstrations $\tau$ produced
by external LLMs and train student with the same OPSD objective.
We consider three demonstration
sources of progressively higher quality: \textbf{MiniMax-M2.7 \footnote{https://huggingface.co/MiniMaxAI/MiniMax-M2.7}}, \textbf{DeepSeekv4-Pro}~\cite{deepseekai2026deepseekv4}, 
and \textbf{Oracle}, which obtain as many correct solutions as possible,
simulating access to all reference solutions.
Details and prompt templates are provided in
Appendix~\ref{app:opsd_details}.

\subsection{Main Results}
\label{sec:main-results}

\textbf{\methodname{} consistently outperforms all baselines.}
Across all five benchmarks (Table~\ref{tab:main-results}),
\methodname{} achieves the highest Avg@8 and Pass@8,
outperforming DAPO by 5.86\% and 8.92\% respectively.
The largest improvements appear on multi-turn agentic tasks,
where credit assignment over long horizons is most challenging.
We provide detailed training dynamics in
Appendix~\ref{app:training_dynamics}.

\paragraph{REINFORCE++ exhibits unstable performance across tasks.}
Despite also operating at the token level, REINFORCE++ shows inconsistent
results: competitive on code generation but significantly weaker on
mathematical reasoning, and collapsing on multi-turn
agentic tasks.  We analyze the training dynamics in Appendix~\ref{app:reinforcepp_dynamics}.
This instability highlights that removing the critic sacrifices
robustness across diverse task structures. In contrast, \methodname{}
remains stable across all five benchmarks.

\textbf{External demonstrations do not reliably help.}
The three OPSD variants all stay close to the Qwen3-8B base model,
substantially below the RL baselines. Unexpectedly, Oracle fails to
outperform DeepSeekv4-Pro and even underperforms the base model on LCB,
suggesting that demonstration quantity and quality are not the bottleneck.
These results indicate that OPSD constructed from external demonstrations
provides an unreliable training signal, as we demonstrate in
\S\ref{sec:opsd-fails}. \methodname{} addresses this by drawing the
teacher trajectory from the model's own verifier-approved rollouts,
achieving substantially stronger performance.

\section{Analysis}
In this section, we conduct a comprehensive analysis to answer the following research questions: 
\textbf{RQ1:} Why must the KL signal be a weight rather than a loss? (\S\ref{sec:kl-weight-vs-loss})
\textbf{RQ2:} Which design choices within \methodname{} matter most? (\S\ref{sec:ablation})
\textbf{RQ3:} Do the gains transfer out of domain? (\S\ref{sec:ood})
\textbf{RQ4:} How does \methodname{} compare with alternative
token-level credit-assignment methods under controlled settings? (\S\ref{sec:controlled-comparison})

\subsection{RQ1: Additive Loss}
\label{sec:kl-weight-vs-loss}
\begin{table}[t]
  \centering
  \small
  \setlength{\tabcolsep}{4pt}
  \renewcommand{\arraystretch}{1.15}
  \begin{tabular}{lcc}
  \toprule
  \textbf{Variant} & \textbf{Avg@8} & \textbf{Pass@8} \\
  \midrule
   GRPO                                  & 39.59 & 43.51 \\
  \methodname{} \textit{[ours]}                  & \textbf{43.22} & \textbf{48.85} \\
  \midrule
  \multicolumn{3}{l}{\textbf{1. Forward KL loss}} \\
  \quad $\beta = 0.1$                   & 38.35 & 41.98 \\
  \quad $\beta : 0.03 \!\to\! 0.1$      & 36.35 & 38.16 \\
  \multicolumn{3}{l}{\textbf{2. Reverse KL loss}} \\
  \quad $\beta = 0.1$                   & 36.06 & 38.93 \\
  \quad $\beta : 0.03 \!\to\! 0.1$      & 39.11 & 41.98 \\
  \multicolumn{3}{l}{\textbf{3. Bidirectional KL loss}} \\
  \quad $\beta_{\mathrm{fwd}}{=}0.1{-}\beta_{\mathrm{rev}}$,\;
        $\beta_{\mathrm{rev}}{:}\,0.03{\to}0.1$
                                        & 34.35 & 36.64 \\
  \bottomrule
  \end{tabular}
  \caption{\textbf{Using additive distillation loss results on LiveCodeBench~v6.}
  All variants use the same self-conditioned
  teacher and optimize
  $\mathcal{L}_{\mathrm{GRPO}} + \beta\,\mathcal{L}_{\mathrm{distill}}$.}
  \label{tab:distill-compare}
  \vspace{-3mm}
\end{table}


A natural alternative to \methodname{} is to add the self-conditioned teacher's KL divergence as an auxiliary distillation loss: $\mathcal{L}_{\mathrm{GRPO}} +
\beta\,\mathcal{L}_{\mathrm{distill}}$. We test Forward KL, Reverse KL (both with fixed $\beta = 0.1$ and linear warm-up $\beta: 0.03 \!\to\! 0.1$), and Bidirectional KL
(combining both directions with complementary coefficients). Table~\ref{tab:distill-compare} shows that all five variants underperform \methodname{}.

\begin{figure}[t]
    \centering
    \includegraphics[width=\linewidth]{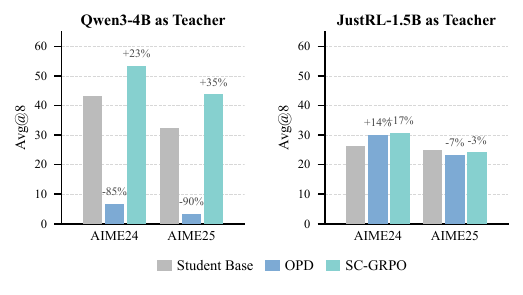}
    \vspace{-7mm}
    \caption{\textbf{Comparison with OPD on AIME.} 
    We compare \methodname{} against OPD using two student models
    on AIME 24 \& 25. 
    Percentages show relative gain over the student baseline.}
    \label{fig:opd}
    \vspace{-2mm}
\end{figure}


Following \citet{li2026rethinkingOPD}, we compare \methodname{} against OPD under two configurations (Figure~\ref{fig:opd}). The left panel (Qwen3-1.7B-Base student,
Qwen3-4B-Base teacher) represents a failing OPD setting, the right panel (DeepSeek-R1-Distill-Qwen-1.5B \citep{shao2024deepseekmath} student, JustRL-DeepSeek-1.5B  \citep{he2025justrl} teacher) represents a successful
setting. Hyperparameters are detailed in Appendix~\ref{sec:opd-hyperparams}. Figure~\ref{fig:opd} shows that \methodname{} consistently outperforms OPD in both settings. When OPD fails (left), \methodname{} achieves 23\% improvement, when OPD
succeeds (right), \methodname{} still outperforms OPD on both AIME tasks, while requiring no external teacher.



\subsection{RQ2: Ablation Study}
\label{sec:ablation}

We validate the three core design choices in \methodname{}
(Table~\ref{tab:ablation}).

\paragraph{A1 Group Routing} The KL weighting is designed to operate where meaningful contrast exists. Partial-solve groups ($2 \le n_c < G$) provide such contrast and benefit from KL weighting. Adding
solve-none groups ($n_c = 0$) brings further improvement, as the diversity signal recovers useful updates from otherwise zero-gradient groups. Including single-correct
groups ($n_c = 1$) violates the design premise: the single verified trajectory dominates the KL signal, causing entropy collapse. Including all-correct groups ($n_c = G$)
similarly fails: without incorrect rollouts for contrast, self-conditioned teacher cannot separate critical decision points from routine tokens. Covering all groups combines both failure modes
and drops below GRPO.


\paragraph{A2 Normalization threshold} The threshold $c$ separates the informative high-KL tail from the uninformative majority (visualization in Appendix~\ref{app:kl_weighting}). 
When $c = p_{50}$ (median), the heavily right-skewed KL distribution pushes $c$
below $10^{-4}$, making $f(\mathrm{KL}) \approx 1$ for nearly all tokens and reducing to uniform weighting. Using $c = \max(p_{75}, 10^{-4})$ achieves the intended
separation: $p_{75}$ suppresses the low-KL majority while the floor prevents collapse when the entire batch has near-zero KL. A fixed constant achieves comparable
performance but requires per-task calibration; $p_{75}$ with floor is adaptive and hyperparameter-free.

\begin{table}[t]
  \centering
  \small
  \setlength{\tabcolsep}{3pt}
  \renewcommand{\arraystretch}{1.2}
  \resizebox{\columnwidth}{!}{%
  \begin{tabular}{lcc}
  \toprule
  \textbf{Variant} & \textbf{Avg@8} & \textbf{Pass@8} \\
  \midrule
  \multicolumn{3}{l}{\textbf{A1. Which groups receive KL weighting?}} \\
  \quad Partial-solve ($2{\le}n_c{\le}7$)
                                             & 41.41 & 45.80 \\
  \quad Partial + solve-none ($\{0\}{\cup}[2,7]$) \textit{[ours]}
                                             & \textbf{43.22} & \textbf{48.85} \\
  \quad Any partial ($1{\le}n_c{\le}7$)
                                             & 38.55 & 41.98 \\
  \quad All-correct only ($n_c{=}8$)
                                             & 40.17 & 42.74 \\
  \quad All groups
                                             & 36.64 & 42.75 \\
  \midrule
  \multicolumn{3}{l}{\textbf{A2. Threshold $c$ $f(\mathrm{KL}) = \mathrm{KL}/(\mathrm{KL}+c)$}} \\
  \quad Adaptive: $\max(p_{75},\, 10^{-4})$ \textit{[ours]}
                                             & \textbf{43.22} & \textbf{48.85} \\
  \quad Fixed: $c = 10^{-4}$                 & 42.93 & 48.09 \\
  \quad Adaptive: $p_{50}$                   & 41.98 & 45.03 \\
  \midrule
  \multicolumn{3}{l}{\textbf{A3. Diversity coefficient $\alpha$}} \\
  \quad $\alpha = 0.1$ \textit{[ours]}       & \textbf{43.22} & \textbf{48.85} \\
  \quad $\alpha = 0.2$                       & 40.74 & 45.80 \\
  \bottomrule
  \end{tabular}
  }%
  \vspace{-2mm}
  \caption{\textbf{Ablation study on LiveCodeBench~v6.} $n_c$ = number of correct rollouts. 
  $p_{50}$/$p_{75}$ denote the 50th/75th percentile of token-level KL values in the current batch. }  
  \label{tab:ablation}
  \vspace{-4mm}
\end{table}

\paragraph{A3 Diversity coefficient}
The diversity signal rewards exploration rather than correctness, so it should serve as a weak exploration signal that does not override the reward gradient. The ablation confirms this: $\alpha{=}0.1$
provides sufficient pressure to break repeated failure patterns, while $\alpha{=}0.2$ lets the exploration signal compete with the RL objective and degrades performance.

\subsection{RQ3: Out-of-Domain Performance}
\label{sec:ood}

\begin{figure}[t]
    \centering
    \includegraphics[width=\linewidth]{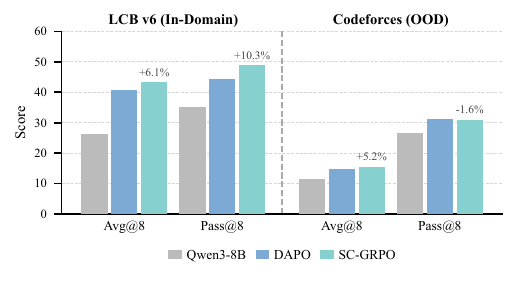}
    \vspace{-7mm}
    \caption{\textbf{In-Domain vs.\ OOD performance.}
    Models trained on LiveCodeBench are evaluated on Codeforces.}
    \label{fig:ood}
    \vspace{-3mm}
\end{figure}

Both \methodname{} and DAPO are trained exclusively on LiveCodeBench and evaluated on Codeforces without domain-specific fine-tuning. Figure~\ref{fig:ood} shows that
\methodname{} retains its advantage, achieving 5.2\% relative gain on Avg@8 (nearly identical to the 6.1\% in-domain gain on LCB~v6) while maintaining comparable Pass@8. The
consistent improvement across domains suggests that \methodname{}'s gains stem from improved credit assignment rather than overfitting.

\subsection{RQ4: Alternative Token-Level Credit Assignment}
\label{sec:controlled-comparison}

To complement the cross-task results in
Table~\ref{tab:main-results}, we compare \methodname{} with recent
credit-assignment methods on LiveCodeBench~v6 under matched settings.
All methods use the same Qwen3-8B base model, training data, verifier,
rollout budget, optimization steps, full-parameter optimization, and
evaluation protocol.

\begin{table}[t]
  \centering
  \small
  \setlength{\tabcolsep}{3.5pt}
  \renewcommand{\arraystretch}{1.15}
  \resizebox{\columnwidth}{!}{%
  \begin{tabular}{llcc}
  \toprule
  \textbf{Method}
  & \textbf{Guidance Source}
  & \textbf{Avg@8}
  & \textbf{Pass@8} \\
  \midrule

  \multicolumn{4}{l}{\textbf{\textit{Alternative credit-assignment methods}}} \\
  SDPO$^\dagger$
  & Verified self-rollout
  & 41.51
  & 45.04 \\
  UCAS
  & Policy uncertainty
  & 40.17
  & 42.75 \\

  \midrule
  \multicolumn{4}{l}{\textbf{\textit{Full-parameter OPSD variants}}} \\
  OPSD
  & MiniMax-M2.7
  & 22.90
  & 35.11 \\
  OPSD
  & DeepSeekv4-Pro
  & 16.79
  & 29.01 \\
  OPSD
  & Oracle
  & 19.85
  & 32.44 \\

  \midrule
  \methodname{} \textit{[ours]}
  & Verified self-rollout
  & \textbf{43.22}
  & \textbf{48.85} \\

  \bottomrule
  \end{tabular}
  }
  \caption{\textbf{Comparison under matched settings on
  LiveCodeBench~v6.} $^\dagger$ For SDPO, the self-teacher is conditioned only on self rollouts from the same problem; no rich environment feedback is provided.}
  \label{tab:matched-comparison}
  \vspace{-3mm}
\end{table}

In table~\ref{tab:matched-comparison}, under matched training and evaluation settings,
\methodname{} outperforms SDPO~\citep{hubotter2025sdpo} and
UCAS~\citep{xie2025unlocking} on both metrics. For SDPO, we restrict the
feedback to verifier-approved self-rollouts, without runtime errors or
failed-test information, thereby matching the supervision available to
\methodname{}. We also evaluate OPSD~\citep{zhao2025opsd} with
full-parameter optimization, instead of the LoRA setting used in
Table~\ref{tab:main-results}. Full-parameter optimization does not improve any of the three OPSD
variants over their LoRA-based counterparts in
Table~\ref{tab:main-results}. Moreover, even the strongest
full-parameter variant, obtaining 22.90 Avg@8 and 35.11 Pass@8,
remains substantially below \methodname{}. These results show that the advantage persists when
both supervision access and optimization capacity are matched.

\section{Conclusion}

We propose \methodname{}, a token-level credit assignment method for RLVR.
By constructing a self-conditioned teacher from the model's own verified rollouts and measuring token-level KL divergence, \methodname{} assigns fine-grained credit to each token.
This KL signal is used as a per-token weight on the GRPO gradient, eliminating sensitivity to divergence form choice and requiring no external resources beyond what RLVR already provides.
Experiments show that \methodname{} consistently outperforms GRPO and other baselines with minimal computational overhead, demonstrating the effectiveness and robustness of \methodname{}.

\subsection*{Limitations}

Due to computational resource constraints, our experiments are limited to models up to 8B parameters and response lengths up to 12288 tokens.
The effectiveness of \methodname{} on larger models and longer responses remains to be explored.
Similarly, resource limitations restrict our evaluation to models operating in standard reasoning mode.
The applicability of \methodname{} to extended thinking modes (e.g., chain-of-thought with explicit reasoning steps, multi-turn refinement) or structured output formats remains unexplored.


\subsection*{Ethics Statement}

This work uses publicly available models, frameworks, and datasets in
accordance with their respective licenses. The solve-none signal
is used only to encourage exploration, not to provide correctness
supervision. Stronger code-generation and agentic policies may introduce existing
risks, including unsafe code generation, tool misuse, and unintended
actions in interactive environments. Models trained with \methodname{}
should therefore be deployed with appropriate sandbox, restricted
tool permissions, and human oversight.

\subsection*{Acknowledgment}
We sincerely thank all the anonymous reviewers for their insightful and detailed feedback. This work is supported by the National Natural Science Foundation
of China (Grant No.U21B2009, 62406015).

\bibliography{custom}

\appendix
\clearpage
\section{Derivation for Direct OPSD in RLVR}
\label{app:opsd-derivation}
\setcounter{equation}{0}
\renewcommand{\theequation}{\thesection.\arabic{equation}}
\renewcommand{\theHequation}{\thesection.\arabic{equation}}

This section formalizes why directly adapting OPSD in the standard RLVR 
setting induces a token-level target mismatch. We fix a query $x$, a 
student rollout prefix $o_{i,<t}$, and analyze the induced distribution 
over the next token.

\subsection{Setup}

Following Section~\ref{sec:preliminaries}, for a query $x$ let 
$\{o_i\}_{i=1}^{G} \sim \pi_{\theta_{\mathrm{old}}}(\cdot \mid x)$ be the 
sampled response group, and let $r(x, o_i) \in \{0, 1\}$ be the verifier 
reward. The set of verifier-approved responses is
\begin{equation}
\mathcal{C}(x)
\;:=\;
\{\, o_i : r(x, o_i) = 1,\ i = 1, \ldots, G \,\}.
\end{equation}
Fix a student response $o_i$ and token position $t$, and let the prefix 
state be $s_{i,t} := (x, o_{i,<t})$. The student next-token distribution is
\begin{equation}
\pi_\theta(\cdot \mid s_{i,t}) 
\;=\; 
\pi_\theta(\cdot \mid x, o_{i,<t}).
\end{equation}
For each verifier-approved response $\tau_j \in \mathcal{C}(x)$, direct 
OPSD constructs a self-conditioned teacher by additionally conditioning the 
same model on $\tau_j$:
\begin{equation}
\widetilde{\pi}_\theta(\cdot \mid s_{i,t},\, \tau_j)
\;:=\;
\operatorname{sg}\!\left[
\pi_\theta(\cdot \mid x,\, \tau_j,\, o_{i,<t})
\right].
\label{eq:app-self-teacher}
\end{equation}
Let $\mu_x$ denote a selection distribution over $\mathcal{C}(x)$ that 
specifies which verifier-approved response is used as the privileged 
context (e.g., uniform over $\mathcal{C}(x)$).

The stop-gradient operator $\operatorname{sg}[\cdot]$ indicates that the 
self-conditioned teacher is treated as a fixed target for the current 
update, and gradients do not backpropagate through teacher logits. We 
further assume all distributions are defined over a common finite 
vocabulary $\mathcal{V}$ with shared support, which is satisfied by 
standard softmax LLM outputs.

\subsection{Forward-KL Direct OPSD}
\label{app:forward-kl-opsd}

We use forward KL as a representative distillation 
objective and analyze direct OPSD pointwise at a 
single prefix.

\paragraph{Notation.}
Fix a query $x$, a student rollout $o_i$, and a token 
position $t$, with prefix state 
$s_{i,t} = (x, o_{i,<t})$. Throughout this subsection 
we suppress the dependence on $(i, t)$ and write
\begin{equation}
\begin{aligned}
q     &\;:=\; \pi_\theta(\cdot \mid s_{i,t}), \\
p_j   &\;:=\; \widetilde{\pi}_\theta(\cdot \mid s_{i,t},\, \tau_j),
\quad \tau_j \in \mathcal{C}(x).
\end{aligned}
\end{equation}
Here $q$ is the student next-token distribution and 
each $p_j$ is a self-conditioned teacher distribution 
(treated as a fixed target via stop-gradient). All 
quantities are distributions over the vocabulary 
$\mathcal{V}$, for a distribution $p$ and a token 
$y \in \mathcal{V}$, $p(y) \in [0,1]$ denotes the 
probability assigned to $y$.

The direct OPSD loss at this prefix is
\begin{equation}
\ell_{\mathrm{OPSD}}(q)
\;:=\;
\mathbb{E}_{j \sim \mu_x}
\!\left[
D_{\mathrm{KL}}\!\bigl(p_j \,\|\, q\bigr)
\right],
\label{eq:app-direct-opsd}
\end{equation}
and the $\mu_x$-averaged teacher distribution is the 
per-token weighted average
\begin{equation}
\bar{p}(y) 
\;:=\; 
\sum_{j} \mu_x(j)\, p_j(y),
\qquad y \in \mathcal{V}.
\label{eq:app-mean-teacher}
\end{equation}

Inserting $\bar p(y)$ into the log ratio in 
Eq.~\eqref{eq:app-direct-opsd} and using 
$\mathbb{E}_j[p_j(y)] = \bar p(y)$ yields
\begin{equation}
\begin{aligned}
\ell_{\mathrm{OPSD}}(q)
&= \mathbb{E}_{j \sim \mu_x}
   \sum_{y \in \mathcal{V}}
   p_j(y)\,\log \tfrac{p_j(y)}{q(y)} \\[2pt]
&= \mathbb{E}_{j \sim \mu_x}
   \sum_{y \in \mathcal{V}}
   p_j(y)\,\log \tfrac{p_j(y)}{\bar p(y)} \\
&\quad + 
   \mathbb{E}_{j \sim \mu_x}
   \sum_{y \in \mathcal{V}}
   p_j(y)\,\log \tfrac{\bar p(y)}{q(y)} \\[2pt]
&= \mathbb{E}_{j \sim \mu_x}\,
   D_{\mathrm{KL}}\!\bigl(p_j \,\|\, \bar p\bigr) \\
&\quad + 
   \sum_{y \in \mathcal{V}}
   \bar p(y)\,\log \tfrac{\bar p(y)}{q(y)} \\[2pt]
&= \underbrace{\mathbb{E}_{j \sim \mu_x}\,
     D_{\mathrm{KL}}\!\bigl(p_j \,\|\, \bar p\bigr)
   }_{\mathcal{D}_{\mathrm{disagree}}\,\geq\,0}
   \;+\; 
   D_{\mathrm{KL}}\!\bigl(\bar p \,\|\, q\bigr).
\end{aligned}
\label{eq:app-decomp}
\end{equation}
where the expectation is over $j \sim \mu_x$ and the 
sums are over $y \in \mathcal{V}$. 
The first term $\mathcal{D}_{\mathrm{disagree}} \geq 0$ 
quantifies how much the self-conditioned teachers 
$\{p_j\}$ disagree with each other at this prefix: 
it is zero exactly when all $p_j$ coincide, and grows as they 
spread apart. Equivalently, when $\mu_x$ is uniform, 
$\mathcal{D}_{\mathrm{disagree}}$ is the generalized 
Jensen--Shannon divergence of $\{p_j\}$.

The disagreement term $\mathcal{D}_{\mathrm{disagree}}$ 
measures how much the self-conditioned teachers 
disagree at this prefix, crucially, it is constant in 
$q$. The only $q$-dependent term is 
$D_{\mathrm{KL}}(\bar p \,\|\, q) \geq 0$, with 
equality if and only if $q = \bar p$. Hence the unconstrained 
pointwise minimizer of 
Eq.~\eqref{eq:app-direct-opsd} is
\begin{equation}
q^\star \;=\; \bar p.
\label{eq:app-forward-optimum}
\end{equation}

Direct forward-KL OPSD does not target any 
particular verified trajectory's local continuation, 
it targets the average $\bar p$ of the 
self-conditioned teacher distributions. The 
disagreement term $\mathcal{D}_{\mathrm{disagree}}$ 
shows that different verified trajectories can induce 
genuinely different local continuations, but this 
disagreement does not influence the optimizer because 
it is constant in $q$. In a parametric policy class, 
the update can be viewed as projecting the student 
toward this average, an average that need not 
correspond to any single feasible trajectory.

\subsection{Reverse-KL Variant}
\label{app:reverse-kl-opsd}

The previous analysis is specific to forward KL. We 
now repeat the pointwise analysis for the reverse-KL 
variant of direct OPSD, using the same 
notation ($q, p_j, \mu_x$) as in 
Appendix~\ref{app:forward-kl-opsd}.

The reverse-KL direct OPSD loss at the prefix 
$s_{i,t}$ is
\begin{equation}
\ell_{\mathrm{OPSD}}(q)
\;:=\;
\mathbb{E}_{j \sim \mu_x}
\!\left[
D_{\mathrm{KL}}\!\bigl(q \,\|\, p_j\bigr)
\right].
\label{eq:app-rev-direct-opsd}
\end{equation}

Unfolding the KL by definition and using the 
linearity of expectation in $j$:
\begin{equation}
\begin{aligned}
\ell_{\mathrm{OPSD}}(q)
&= \mathbb{E}_{j}
   \sum_{y \in \mathcal{V}} q(y)\,
   \log \tfrac{q(y)}{p_j(y)} \\[2pt]
&= \sum_{y \in \mathcal{V}} q(y) \log q(y) \\
&\quad - \sum_{y \in \mathcal{V}} q(y)\,
   \mathbb{E}_{j}\!\left[\log p_j(y)\right],
\end{aligned}
\label{eq:app-rev-expand}
\end{equation}

Minimizing Eq.~\eqref{eq:app-rev-expand} over $q$ 
subject to $\sum_y q(y) = 1$, we form the Lagrangian
\begin{equation}
\begin{aligned}
\mathcal{L}(q, \lambda)
&= \sum_{y} q(y) \log q(y) \\
&\quad - \sum_{y} q(y)\,\mathbb{E}_{j}[\log p_j(y)] \\
&\quad + \lambda\!\left(\sum_{y} q(y) - 1\right),
\end{aligned}
\end{equation}
where $\lambda$ is the Lagrange multiplier for the 
normalization constraint. Setting 
$\partial \mathcal{L} / \partial q(y) = 0$ yields 
the stationarity condition
\begin{equation}
\log q(y) + 1 - \mathbb{E}_{j}\!\left[\log p_j(y)\right] + \lambda = 0,
\end{equation}
which gives 
$q(y) \propto \exp\!\bigl(\mathbb{E}_{j}[\log p_j(y)]\bigr)$. 
Enforcing the normalization constraint yields
\begin{equation}
q^\star(y)
\;=\;
\frac{
\exp\!\bigl(\mathbb{E}_{j}[\log p_j(y)]\bigr)
}{
\sum_{y' \in \mathcal{V}}
\exp\!\bigl(\mathbb{E}_{j}[\log p_j(y')]\bigr)
}.
\label{eq:app-reverse-optimum}
\end{equation}
Equivalently, $q^\star$ is the normalized weighted 
geometric mean of the teacher distributions:
\begin{equation}
q^\star(y) 
\;\propto\; 
\prod_{j} p_j(y)^{\mu_x(j)}.
\label{eq:app-reverse-geom}
\end{equation}

\paragraph{Interpretation}
Reverse-KL direct OPSD replaces the per-token 
arithmetic average $\bar p$ from the forward-KL case 
(Eq.~\eqref{eq:app-forward-optimum}) with a 
normalized \emph{geometric} average of the 
self-conditioned teacher distributions. The two 
averages have very different behavior: the geometric 
mean $q^\star(y)$ is large only when \emph{every} 
$p_j(y)$ is non-negligible, so reverse KL exhibits 
mode-seeking behavior, concentrating mass on tokens 
that all teachers find plausible. The arithmetic 
mean $\bar p(y)$, in contrast, is large whenever 
\emph{any} $p_j(y)$ is large, exhibiting 
mode-covering behavior. Despite this difference, the 
overall conclusion is the same: changing the 
divergence merely changes the form of averaging, but 
the optimization target is still constructed from 
self-conditioned teacher distributions.

\subsection{Summary: Why Direct OPSD Fails in RLVR}
\label{app:opsd-summary}

Combining Appendices~\ref{app:forward-kl-opsd} 
and~\ref{app:reverse-kl-opsd}, we obtain a unified 
characterization of direct OPSD at any prefix 
$s_{i,t}$.

\paragraph{Pointwise minimizers under different divergences.}
Let $\{p_j\}_{j}$ denote the self-conditioned teacher 
distributions induced by verifier-approved 
trajectories $\tau_j \in \mathcal{C}(x)$, weighted by 
the selection rule $\mu_x$. Then the pointwise 
minimizer of direct OPSD takes the form
\begin{equation}
q^\star(y) \;=\;
\begin{cases}
\displaystyle \sum_{j} \mu_x(j)\, p_j(y) 
& \text{(forward KL),} \\[6pt]
\displaystyle \frac{\prod_j p_j(y)^{\mu_x(j)}}
                   {\sum_{y'}\prod_j p_j(y')^{\mu_x(j)}}
& \text{(reverse KL),}
\end{cases}
\label{eq:app-summary-optima}
\end{equation}
a weighted arithmetic average and a normalized 
weighted geometric average of the teachers, 
respectively.

\paragraph{Three structural problems.}
Eq.~\eqref{eq:app-summary-optima} exposes three 
problems that arise regardless of the choice of 
divergence:

\textbf{The optimization target is an average, 
not a trajectory.} Both minimizers are constructed 
by combining multiple self-conditioned 
teachers. Even if every individual $p_j$ corresponded 
to a valid local continuation of $\tau_j$, the 
average $q^\star$ generally does not correspond to 
any single feasible trajectory. The student is 
trained to imitate a synthetic distribution that no 
verified rollout actually instantiates.

\textbf{Disagreement between teachers carries 
information, but the loss discards it.} The 
forward-KL decomposition in 
Eq.~\eqref{eq:app-decomp} contains the disagreement 
term $\mathcal{D}_{\mathrm{disagree}} \geq 0$, which 
measures how much the teachers $\{p_j\}$ disagree 
about the next token at the current prefix. This 
quantity is large precisely at prefixes where 
different verified trajectories branch into 
different continuations---for example, a math prefix 
``\texttt{Solve $2x+4=10$, so}'' where one verified 
trajectory continues with ``\texttt{$2x = 6$}'' 
while another continues with ``\texttt{subtract 4 
from both sides}''. Such prefixes are arguably the 
most informative ones for the student: they mark 
decision points where multiple valid continuations 
exist. Yet because $\mathcal{D}_{\mathrm{disagree}}$ 
does not depend on $q$, it contributes nothing to 
the gradient. As far as the loss is concerned, a 
prefix with strong teacher disagreement is treated 
identically to a prefix where all teachers agree. 
Direct OPSD therefore uses the disagreement signal 
only \emph{implicitly}, through how it perturbs the 
average $\bar p$, and never as a per-token quantity 
that could focus learning on these branch points.

\textbf{The teacher is not a verifier-defined 
ground truth.} Each $p_j$ is constructed by 
conditioning the student on a single 
verifier-approved trajectory $\tau_j$. As discussed 
in Section~\ref{sec:method}, $\tau_j$ certifies only 
end-task success and may contain redundant, lucky, 
or even flawed reasoning. Treating $p_j$ as a 
distillation target therefore propagates whatever 
artifacts $\tau_j$ contains into the student, 
weighted equally with genuinely informative 
predictions.

\paragraph{Implication.}
The three problems above are not artifacts of a 
particular divergence choice or selection rule 
$\mu_x$: they follow directly from the structural 
fact that direct OPSD uses self-conditioned teachers 
as \emph{distillation targets}. Any objective that 
asks the student to move toward $\{p_j\}$ in 
distribution will inherit some form of 
Eq.~\eqref{eq:app-summary-optima}, and hence inherit 
(P1)--(P3).

\section{Experiment Details}
\subsection{Training Details}
\label{app:training_details}

Table~\ref{tab:training_settings} summarizes the training settings for each task.
All experiments use \textbf{Qwen3-8B (thinking disabled)} as the base model.
For mathematical reasoning (AIME 24\&25), we randomly sample 4{,}000 problems from DAPO-Math-17k~\citep{yu2026dapo}.
For code generation (LCB v6), we randomly sample half of the unit tests per problem in LiveCodeBench v6 \cite{jain2025livecodebench} as our training set and reserve the other half for evaluation.
For agentic tasks, effective behavior requires long-horizon planning,
coordination of heterogeneous tools or APIs, persistent state tracking
across interactions, and adaptation to changing environments
\citep{wang2024appbench,
cheng2026terminalworldscalingterminalagentenvironments,
liu2026docosproactivedocumentguidedactions}, we choose AppWorld and WebShop.
For AppWorld \cite{trivedi2024appworld}, we train on the official training split.
For WebShop \cite{yao2022webshop}, we randomly sample 2{,}400 tasks from the training split.
All runs use 8 rollouts per prompt, a batch size of 8 prompts per update, and a learning rate of \(1 \times 10^{-6}\).
We conducted experiments using an 8-node cluster. Each node was equipped with 8 NVIDIA A100 80GB GPUs, 144-core AMD EPYC 7713 processors, and 960 GB of RAM, running Ubuntu as the operating system.

\begin{table*}[ht]
\centering
\begin{tabular}{lccccc}
\toprule
\textbf{Task} & \textbf{Training Data} & \textbf{Size} & \textbf{Max Resp. Len} & \textbf{LR} & \textbf{Steps} \\
\midrule
AIME 24\&25 & DAPO-Math-17k    & 4000 & 8192  & 1e-6 & 500 \\
LCB v6      & LiveCodeBench v6 & 131  & 8192  & 1e-6 & 320 \\
AppWorld    & AppWorld         & 90   & 12288 & 1e-6 & 110 \\
WebShop     & WebShop          & 2400 & 8192  & 1e-6 & 300 \\
\bottomrule
\end{tabular}
\caption{Training settings for each task. ``Size'' denotes the number of training set size. ``Max Resp. Len'' is the maximum response length in tokens.}
\label{tab:training_settings}
\end{table*}

\subsection{OPSD Details}
\label{app:opsd_details}

We follow the OPSD framework of \citet{zhao2025opsd}. In the
original formulation, the teacher is conditioned on the ground-truth
answer $y^*$ (e.g., an answer with reference chain-of-thought), and the training
objective minimizes the per-token divergence between the privileged
teacher and the student along the student's own rollouts:
\begin{align*}
&\mathcal{L}_{\mathrm{OPSD}}(\theta)
= \mathbb{E}_{x,\,\hat{y}\sim\pi_\theta(\cdot\mid x)}
\left[
\frac{1}{|\hat{y}|}\sum_{t=1}^{|\hat{y}|}
D\big(p_T^t\,\big\|\,p_S^t\big)
\right]\\[4pt]
&\text{where}\quad
p_T^t = \pi_\theta(\cdot \mid x, y^*, \hat{y}_{<t}),\\
&\phantom{\text{where}}\quad
p_S^t = \pi_\theta(\cdot \mid x, \hat{y}_{<t}).
\end{align*}

\paragraph{Demonstration Acquisition}
The key requirement of OPSD is a ground-truth
answer $y^*$ to serve as the teacher's privileged context. How $y^*$ is obtained
depends on whether ground-truth answers are available.

\textbf{Math (DAPO-Math-17k).} Ground-truth answers are
  available, so we directly use them as the privileged context
  $y^*$ to construct the self-conditioned teacher. To simulate
  varying teacher quality, we sample 500 questions and measure each
  external LLM's accuracy; we then provide ground-truth
  to a corresponding fraction of training queries
  (matching that model's coverage rate). Oracle provides ground
  truth to 100\% of queries.

\textbf{Code generation (LCB~v6), AppWorld, and WebShop.} No
  ground-truth solutions exist. We instead query each teacher LLM
  up to 3 times per query and retain any correct trajectory
  (verified by execution) as the demonstration $\tau$. Per-model
  coverage is reported in Table~\ref{tab:opsd-coverage}.

\begin{table}[t]
  \centering
  \small
  \setlength{\tabcolsep}{4pt}
  \renewcommand{\arraystretch}{1.15}
  \resizebox{\columnwidth}{!}{%
  \begin{tabular}{lcccc}
  \toprule
  \textbf{Teacher Source}
    & \textbf{DAPO-Math}
    & \textbf{LCB v6}
    & \textbf{WebShop}
    & \textbf{AppWorld} \\
  \midrule
  MiniMax-M2.7       & 93.60 & 46.56 & 43.38 & 70.00 \\
  DeepSeekv4-Pro     & 97.80 & 67.94 & 50.21 & 78.89 \\
  Oracle             & 100.00  & 79.39 & 64.13 & 86.67 \\
  \bottomrule
  \end{tabular}
  }%
  \caption{\textbf{OPSD teacher coverage (\%, 3 attempts).} For each query, we sample up to
  three attempts and retain any correct trajectory as the
  demonstration. Math uses ground-truth answer verification; code
  and agentic tasks use execution-based verifiers.}
  \label{tab:opsd-coverage}
\end{table}

\paragraph{Training Hyperparameters}
All OPSD experiments use Qwen3-8B as the base model and train LoRA adapters with bfloat16 precision. Following \citep{zhao2025opsd}, we use LoRA
rank $r=64$ and scaling factor $\alpha=128$ on all attention and MLP projection layers
(\texttt{q\_proj}, \texttt{k\_proj}, \texttt{v\_proj}, \texttt{o\_proj}, \texttt{gate\_proj}, \texttt{up\_proj}, and \texttt{down\_proj}).
We optimize with learning rate $1\times 10^{-6}$, gradient clipping at 1.0, per-device batch size 1, gradient accumulation 2, and 8
training processes, yielding an effective batch size of 16.
For OPSD, we use a fixed teacher, $\beta=0$, $\lambda=1$, temperature 1.0, top-$p=0.95$, and token-level JSD clipping threshold 0.05.

\paragraph{OPSD Prompt}
\label{app:opsd-prompt-template}
The student prompt for each task follows that used in
our main experiments. The teacher
prompt injects a reference/answer block as privileged context. Below we
show the teacher-specific prompt for each task type.

\begin{promptbox}[Math Teacher Prompt]
\begin{lstlisting}[style=promptstyle]
Problem: {problem}

(*\textcolor{phaseblue}{\textbf{The correct answer to this problem is: \textbackslash boxed\{\{answer\}\}.}}*)

Now derive a complete step-by-step solution that arrives at this answer. Think carefully and show all reasoning. Put your final answer within \boxed{}.
\end{lstlisting}
\end{promptbox}

\begin{promptbox}[Code (LCB) Teacher Prompt]
\begin{lstlisting}[style=promptstyle]
{Standard LiveCodeBench SYSTEM PROMPT}

(*\textcolor{phaseblue}{\textbf{Here is a reference solution that passes all test cases:}}*)
```python
(*\textcolor{phaseblue}{\textbf{\{reference solution\}}}*)
```
Now solve the problem yourself. Think step by step, write clean Python code, and put your final solution in a ```python``` code block.
\end{lstlisting}
\end{promptbox}

\begin{promptbox}[WebShop Teacher Prompt]
\begin{lstlisting}[style=promptstyle]
{Standard WebShop SYSTEM PROMPT}

(*\textcolor{phaseblue}{\textbf{Here is a reference trajectory that successfully completed this task:}}*)
(*\textcolor{phaseblue}{\textbf{\{successful trajectory\}}}*)

Now complete this shopping task yourself, one action at a time.
\end{lstlisting}
\end{promptbox}

\smallskip\noindent

In all cases, the reference block is appended to the standard
task prompt. The student never sees the reference block.

\subsection{OPD Details}
\label{sec:opd-hyperparams}
\begin{table}[t]
\centering
\small
\begin{tabular}{lcc}
\toprule
Hyperparameter & Training & Inference \\
\midrule
Max prompt length & 4096 & 1024 \\
Max response length & 7168 & 31744 \\
Thinking Mode & False & True \\
\bottomrule
\end{tabular}
\caption{Training and inference hyperparameters used for the OPD comparison experiment (Figure~\ref{fig:opd}). \label{tab:opd-hyperparams}}
\end{table}

To align with the settings in \cite{li2026rethinkingOPD}, we use the same training and inference configurations as \cite{li2026rethinkingOPD} in Figure~\ref{fig:opd}. See Table~\ref{tab:opd-hyperparams} for details. 
Due to computational resource constraints, we sample 3,600 examples for RL training, which corresponds to the 60 step performance reported in \cite{li2026rethinkingOPD}.

\subsection{Analysis of Reinforce++}
\label{app:reinforcepp_dynamics}

REINFORCE++~\citep{hu2025reinforcepp} applies per-token credit assignment
through discounted returns: $G_t = r_t + \gamma G_{t+1}$. While competitive
on short-horizon tasks (math, code), it collapses on long-horizon agentic tasks.

On AppWorld, validation performance peaked at 0.175@step40, then declined to
0.035@step101. On WebShop, performance dropped from 0.327 (base model) to 0.0@step80,
with training reward also reaching 0.0 in the final steps.

The instability stems from gradient variance in long horizons. Early tokens
accumulate returns from all subsequent tokens: $G_1 = \sum_{t=1}^T \gamma^{t-1} r_t$.
With $\gamma=1.0$ and trajectories spanning hundreds of tokens, gradient variance
grows exponentially. Combined with sparse episode-level rewards, advantage whitening
fails to identify critical tokens, causing policy collapse.

In contrast, \methodname{} uses trajectory-level GRPO advantages and
applies token-level credit assignment only through KL-based gradient modulation,
avoiding variance explosion in long-horizon settings.

\section{Training Dynamics}
\label{app:training_dynamics}

\subsection{Training Dynamics on AIME 24 \& 25}
\label{app:aime_training_curves}

\begin{figure*}[ht]
    \centering
    \includegraphics[width=\linewidth]{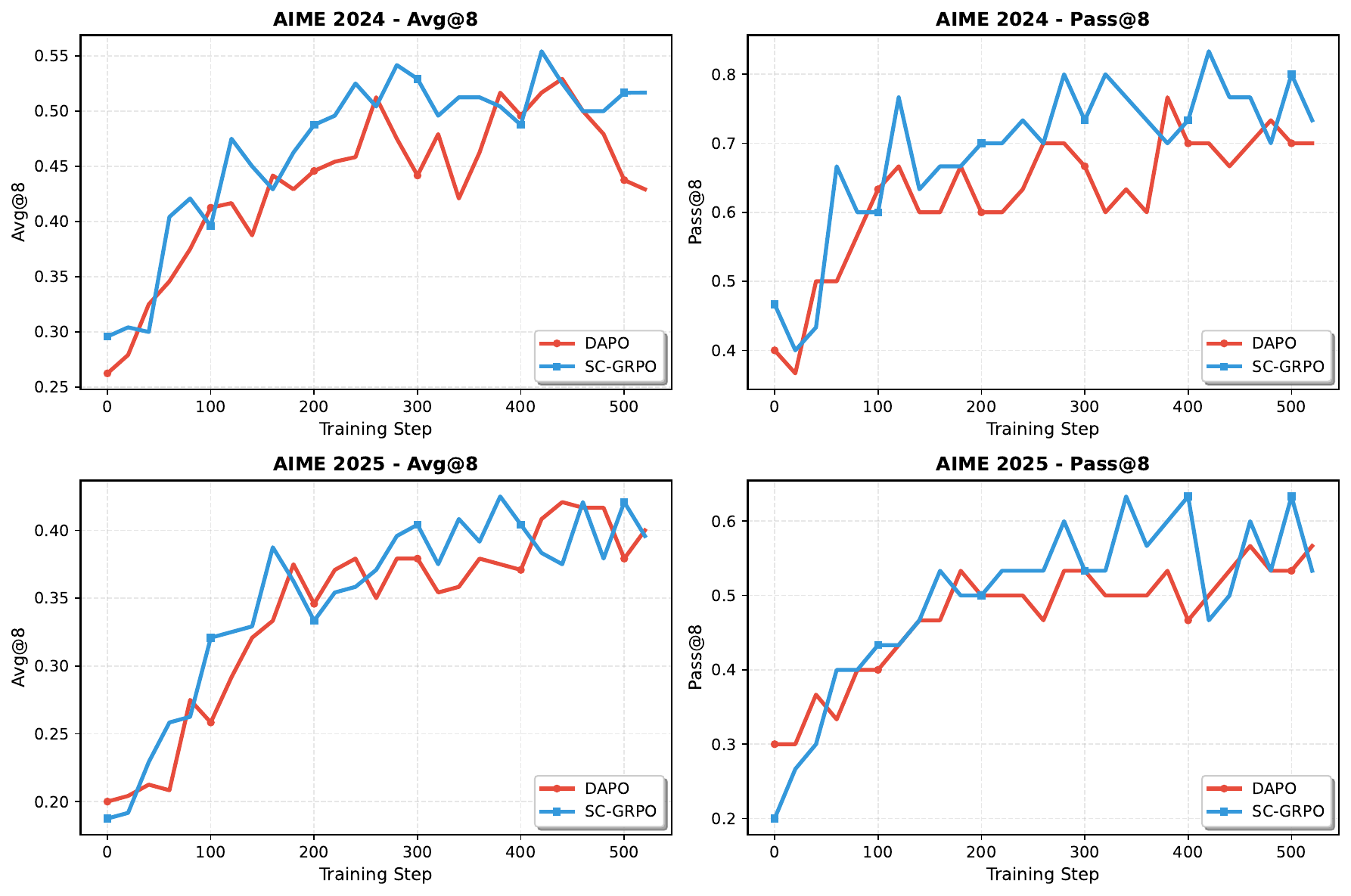}
    \caption{\textbf{Training dynamics on AIME 24 \& 25.}
    We compare validation performance of \methodname{} (blue) and DAPO (red)
    throughout training. \methodname{} consistently outperforms DAPO on both
    Avg@8 and Pass@8 metrics across both test sets.}
    \label{fig:training-dynamics-aime}
\end{figure*}

Figure~\ref{fig:training-dynamics-aime} shows the validation performance
throughout training on AIME 2024 and 2025. Both methods are trained on
DAPO-Math-17k with identical hyperparameters and evaluated every 20 steps.

\paragraph{AIME 2024 (Top Row)}
On Avg@8 (left), \methodname{} establishes an early advantage and maintains
it throughout training.
Both methods exhibit some variance in the later stages of training, but
\methodname{} consistently stays above DAPO. On Pass@8 (right), the gap is
more pronounced: \methodname{} achieves 0.80 at step 280, while DAPO peaks
at 0.72 around step 260 before declining slightly.

\paragraph{AIME 2025 (Bottom Row)}
AIME 2025 is harder than AIME 2024, with both methods achieving
lower absolute scores. On Avg@8 (left), \methodname{} shows steady improvement
throughout training, reaching 0.40 by step 500.
 The advantage is more consistent here than on AIME 2024, with less
variance in the later stages. On Pass@8 (right), \methodname{} maintains
a 5-10\% advantage throughout most of training, reaching
0.63 compared to DAPO's 0.57.

\paragraph{Stability and convergence}
Both methods exhibit training variance typical of RL with verifiable rewards,
where validation performance fluctuates due to the discrete nature of the
reward signal. However, \methodname{} demonstrates more stable improvement
on AIME 2025, the harder test set, suggesting that token-level credit
assignment becomes increasingly valuable as problem difficulty increases.
The consistent gap across both metrics and both test sets confirms that
the gains are not due to overfitting to a particular evaluation protocol.

\subsection{Policy Entropy}
\label{app:policy_entropy}

\begin{figure}[ht]
    \centering
    \includegraphics[width=\linewidth]{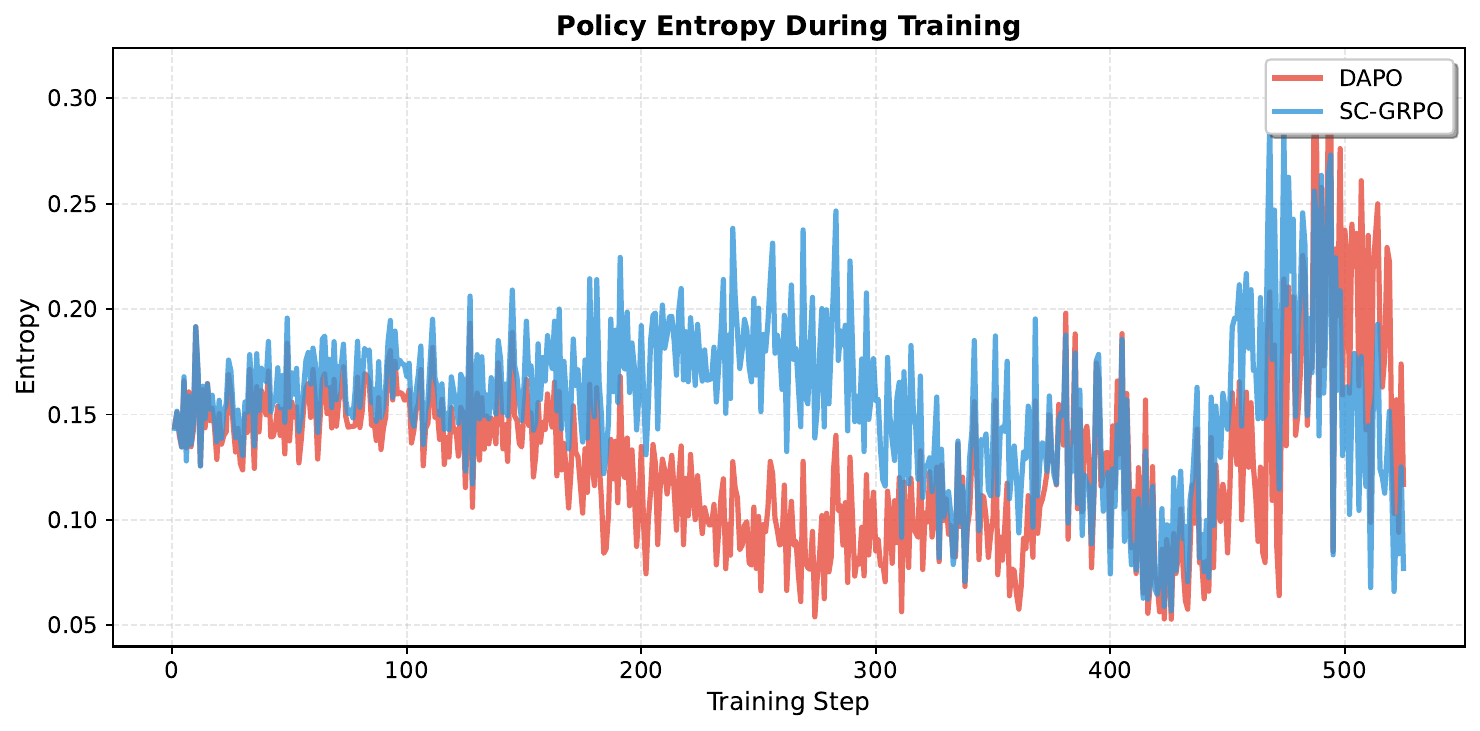}
    \caption{\textbf{Policy entropy during training.}
    We compare the entropy of \methodname{} (blue) and DAPO (red) throughout
    training on DAPO-Math-17k. \methodname{} maintains consistently higher
    entropy, indicating better exploration and diversity preservation.}
    \label{fig:training-entropy}
\end{figure}

Figure~\ref{fig:training-entropy} shows the policy entropy throughout training.
\methodname{} maintains consistently higher entropy than DAPO across all
training steps, with the gap widening after step 200. This suggests that
\methodname{} preserves exploration capacity: by selectively
suppressing gradients on tokens where the student already matches the teacher
and encouraging exploration when facing solve-none groups, \methodname{} avoids premature convergence on the full trajectory.
In contrast, DAPO's uniform credit assignment drives the policy toward
deterministic outputs more aggressively, reducing entropy and potentially
limiting the model's ability to explore alternative reasoning paths.

The entropy gap correlates with the performance advantage observed in
Figure~\ref{fig:training-dynamics-aime}: higher entropy enables the model
to maintain diverse solution strategies, which is particularly valuable
on harder problems (AIME 2025) where multiple reasoning approaches may
be necessary to reach the correct answer.

\subsection{KL Weighting Function}
\label{app:kl_weighting}

\begin{figure}[ht]
    \centering
    \includegraphics[width=\linewidth]{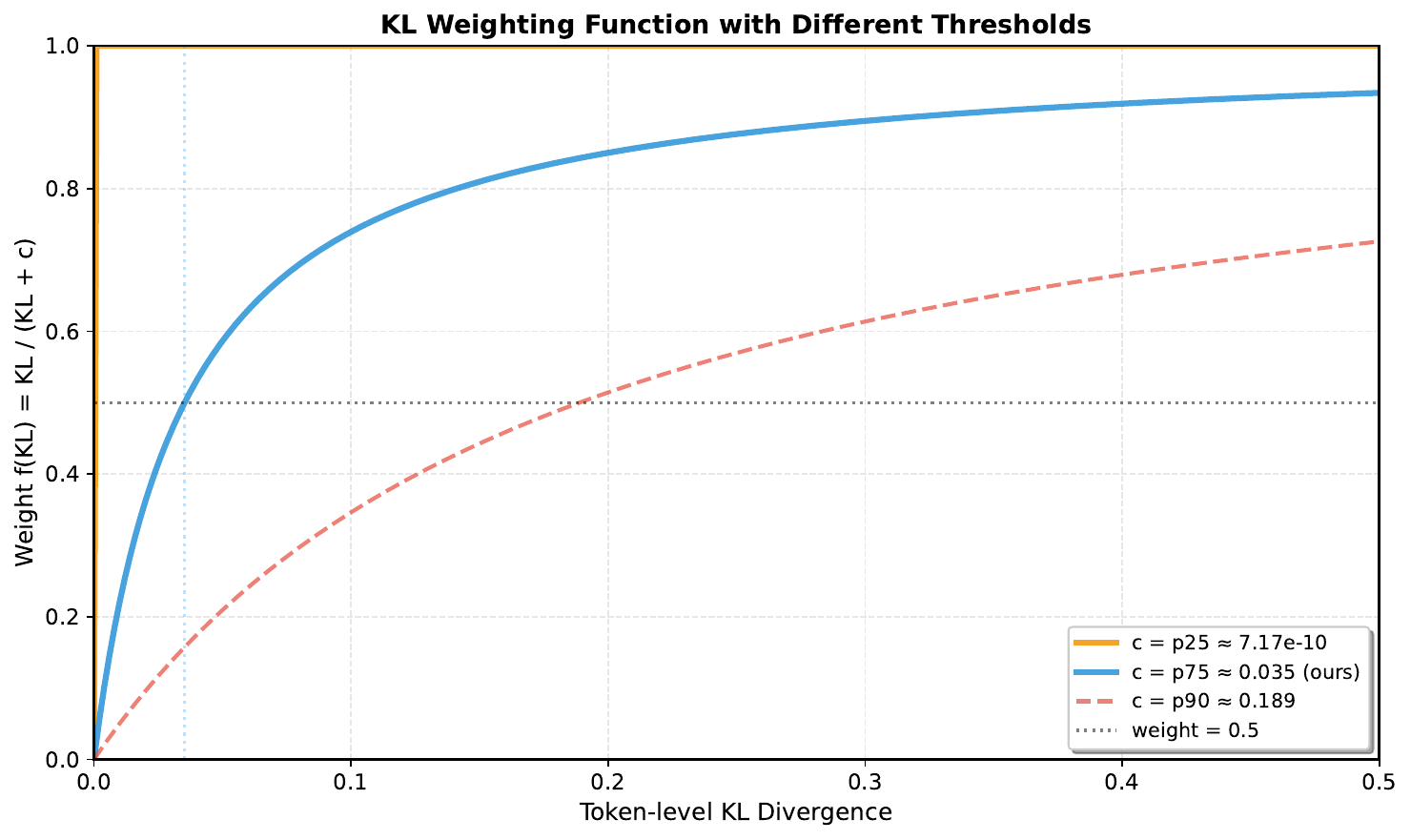}
    \caption{\textbf{KL weighting function with different thresholds.}
    We visualize the weighting function $f(\text{KL}) = \text{KL} / (\text{KL} + c)$
    for different choices of threshold $c$ \textbf{at training step 270}. Using $c = p_{75}$ (blue) effectively
    separates tokens where the student matches the teacher (KL $< c$, weight $< 0.5$)
    from critical difference tokens (KL $> c$, weight $> 0.5$). Using $c = p_{25}$
    (orange) provides almost no discrimination, while $c = p_{90}$ (red) over-suppresses
    moderately divergent tokens.}
    \label{fig:kl-weighting}
\end{figure}

Figure~\ref{fig:kl-weighting} illustrates why we use the 75th percentile
of token-level KL divergence as the threshold $c$ in our weighting function.
The function $f(\text{KL}) = \text{KL} / (\text{KL} + c)$ maps each token's
KL divergence to a weight between 0 and 1, which modulates the RL gradient
for that token.

When $c = p_{25}$ (orange curve), the threshold is near zero, so almost all
tokens receive high weights ($f(\text{KL}) \approx 1$), effectively reverting
to uniform credit assignment. When $c = p_{90}$ (red curve), the threshold
is too high, causing many tokens with moderate KL divergence (0.05--0.15)
to receive low weights, over-suppressing the gradient signal.

Our choice of $c = p_{75}$ (blue curve) strikes a balance: tokens with
KL below $p_{75}$ (the bottom 75\% of the distribution) receive weights
below 0.5, indicating the student has already learned these tokens from
the teacher. Only the top 25\% of tokens, those with the highest KL
divergence, receive weights above 0.5, preserving the RL gradient where
the student and teacher genuinely differ. This adaptive threshold ensures
that credit assignment focuses on critical difference tokens while avoiding
both under-discrimination and over-suppression.

\subsection{KL Distribution Evolution}
\label{app:kl_distribution}

\begin{figure}[ht]
    \centering
    \includegraphics[width=\linewidth]{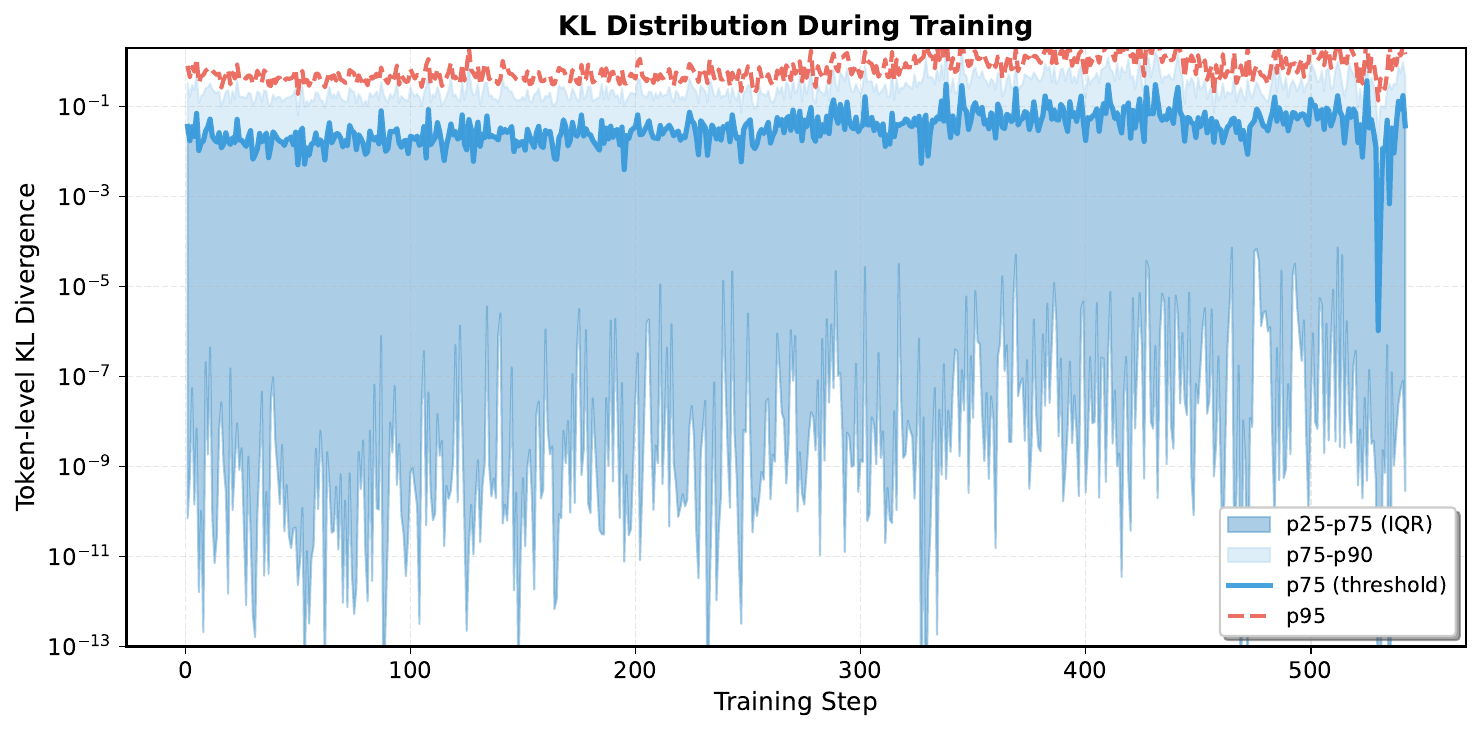}
    \caption{\textbf{Evolution of KL distribution during training.}
    We track the percentiles of token-level KL divergence throughout training.
    The shaded regions show the interquartile range (p25-p75, dark blue) and
    the p75-p90 range (light blue). The p75 line (blue) serves as our adaptive
    threshold, while p95 (red dashed) marks the upper tail. The distribution
    remains stable throughout training, with p75 consistently separating the
    bottom 75\% of tokens (low KL, student matches teacher) from the top 25\%
    (high KL, critical difference tokens).}
    \label{fig:kl-distribution}
\end{figure}

Figure~\ref{fig:kl-distribution} shows how the KL distribution evolves
throughout training. The distribution exhibits a heavy-tailed structure:
the bottom 75\% of tokens (p25 to p75, dark blue region) have very low
KL divergence, often near zero, indicating that the student policy has
already learned to match the teacher on these tokens. Only the top 25\%
of tokens show significant divergence.

This distribution remains remarkably stable across training steps, with
p75 fluctuating around 0.03--0.10 and p95 around 0.2--0.5. The stability
validates our choice of using p75 as an adaptive threshold: it consistently
identifies the boundary between learned tokens and critical difference tokens,
regardless of training stage. The occasional spikes in the lower percentiles
(visible as vertical streaks in the p25-p75 region) correspond to batches
with particularly challenging problems where even common tokens require
learning, but these are rare and do not affect the overall threshold selection.

\section{Case Study: Token-Level KL Heatmap}
\label{app:case-study}

\begin{figure*}[t]
  \centering
  \includegraphics[width=0.85\textwidth]{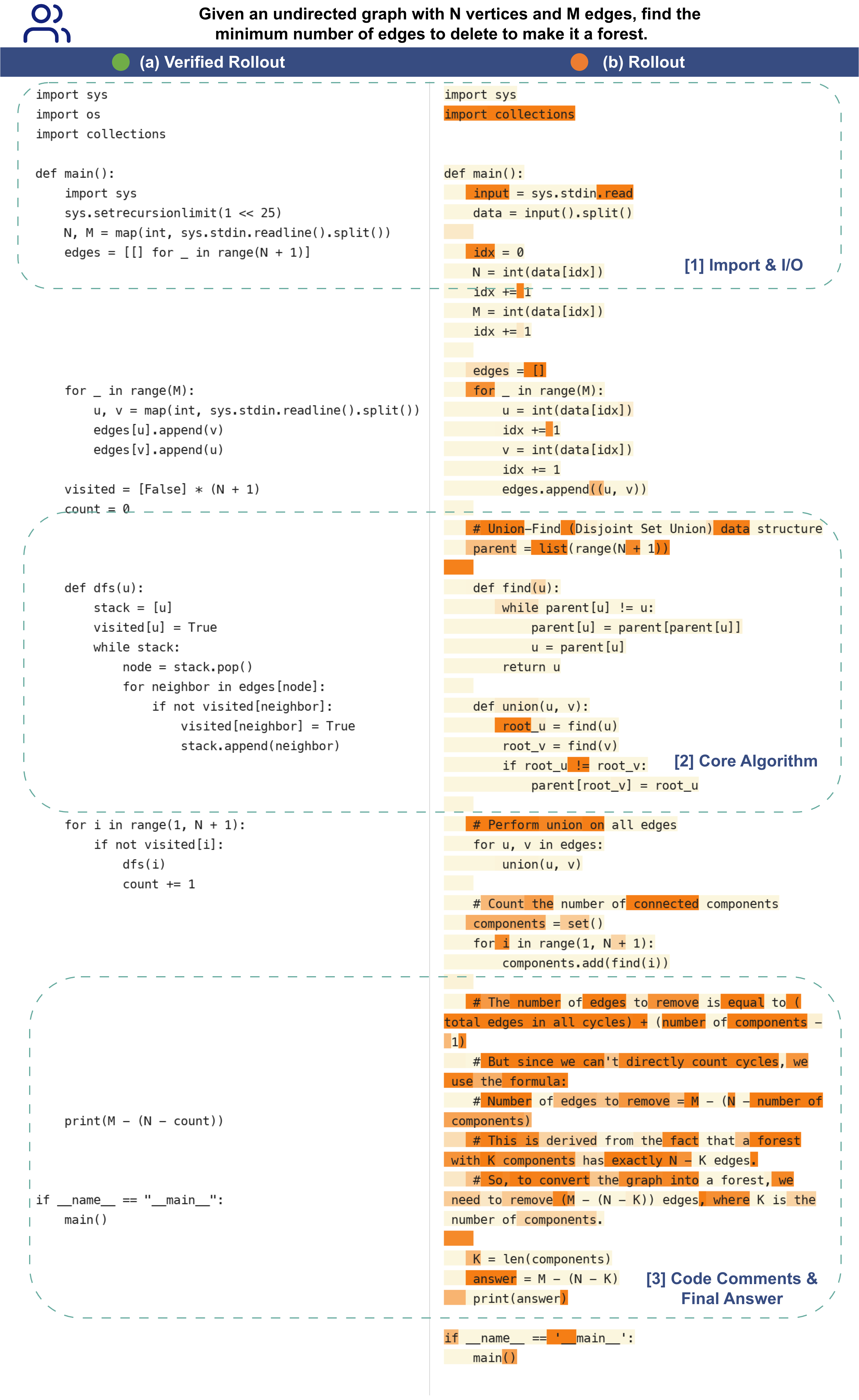}
  \caption{\textbf{Token-level KL heatmap on a LiveCodeBench rollout.}
  Each rollout token is shaded by its KL weight $f_t$, which measures the
  shift in the student's next-token distribution induced by inserting the
  verified trajectory $\tau$ into the teacher's context.}
  \label{fig:case-study}
\end{figure*}

Figure~\ref{fig:case-study} shows the per-token KL distribution on a real
LiveCodeBench problem \textit{(counting the minimum number of edges to remove so that
an undirected graph becomes a forest)}. The student rollout $o_i$ implements
the standard \emph{Disjoint Set Union} (DSU / Union-Find) approach, while the 
sampled verifier-approved trajectory~$\tau$ used to condition the teacher implements a
stack-based \emph{Depth-First Search} (DFS) over an adjacency list. Both
solve the problem correctly, but follow distinct algorithmic paths.

Panels~(a) show the sampled verifier-approved trajectory~$\tau$
that is injected into the teacher's system prompt. Panel~(b) shows the
student rollout $o_i$, with each token shaded by its KL weight
$f_t = D_t / (D_t + c)$. Near-white tokens correspond to positions where the
student's distribution is essentially unchanged by conditioning on $\tau$:
the model would produce these tokens whether or not a verified
solution was available. Deeply shaded tokens mark positions where $\tau$
substantially redirects the next-token distribution, these are the tokens
on which the GRPO update is preserved at full weight, while the
near-white tokens are downweighted.

\paragraph{What tokens are selected}
Most low-level scaffolding code (indentation, parentheses, simple variable
assignments, generic loops) receives near-zero weight,
since both DSU and DFS share these surface forms. The high-KL tokens
concentrate at the points where the two algorithms genuinely diverge:
the import of \texttt{collections} and the choice of stdin reader; the
allocation of the DSU \texttt{parent = list(range(N+1))} array
(absent under DFS, which builds an adjacency list instead); the bodies of
\texttt{def find} and \texttt{def union}, the merge predicate
\texttt{root\_u != root\_v}, and the construction of the
\texttt{components} set; and finally the closed-form answer
\texttt{answer = M - (N - K)}. These are precisely the tokens whose choice
distinguishes a DSU implementation from a DFS one, and on which a verified
DFS trajectory shifts the student's belief most strongly.

The pattern is consistent with the design intent of \methodname: the self-teacher
does not act as an additional supervision signal, it acts as a
\emph{credit-assignment filter} on the existing GRPO gradient. Sequence-level
reward is preserved as the source of update direction, while gradient mass
is concentrated on the small subset of tokens that actually carry the
algorithmic decision.

\section{Prompt Templates for \methodname}
\label{app:prompts}

\subsection{Prompt Construction}
\label{app:prompt-construction}

For a given problem instance, we construct \textbf{student} and \textbf{teacher} prompts as follows:
The student prompt consists of a base system prompt followed by the problem specification. The teacher prompt extends this by concatenating three components: the base system prompt, the guide instruction, and the reference trajectory.

\subsection{LiveCodeBench Templates Example}
\label{app:prompt-lcb}

We use LiveCodeBench as an example to illustrate our prompt templates. The same structure applies to other tasks (math and agentic tasks) with domain-specific guide instructions.

\begin{promptbox}[Base System Prompt]
\begin{lstlisting}[style=promptstyle]
You are an expert Python programmer. You will be given a question (problem
specification) and will generate a correct Python program that matches the
specification and passes all tests.
\end{lstlisting}
\end{promptbox}

\begin{promptbox}[Guide Prompt]
\begin{lstlisting}[style=promptstyle]
A reference correct solution is provided below. Use it as a high-level
algorithmic guide, not as code to copy. Your goal is to write a solution
that is correct, clean, and efficient:
- Correct: passes all test cases within the time limit.
- Clean: well-structured, readable, and free of dead code or redundant logic.
- Efficient: chooses an algorithm with appropriate time and space complexity.

When using the reference solution:
1. Understand the core algorithm and data structures it uses.
2. Adopt its strategy if it is sound; improve on it if you can.
3. Do not copy variable names, formatting, or boilerplate from the reference.
4. Handle all edge cases independently, do not assume the reference covers them.
5. Write your own solution from scratch, guided by the reference algorithm only.
\end{lstlisting}
\end{promptbox}

\begin{promptbox}[Reference Template]
\begin{lstlisting}[style=promptstyle]
### Reference Correct Solution
{reference_trajectory}

Use the reference algorithm as inspiration only.
Write your own complete, correct Python solution from scratch.
\end{lstlisting}
\end{promptbox}

For math tasks (DAPO-Math-17K) and agentic tasks (AppWorld \& Webshop), we use task-specific system prompts and guide instructions.
\end{document}